%% file: main.tex
\documentclass[twoside]{article}

\usepackage[accepted]{aistats2026}
\usepackage[round]{natbib}

\input{head}

\begin{document}

\runningtitle{Robust Estimation of Heterogeneous Treatment Effects in Randomized Trials Leveraging External Data}


\twocolumn[

\aistatstitle{Robust Estimation of Heterogeneous Treatment Effects\\ in Randomized Trials Leveraging External Data}

\aistatsauthor{Rickard Karlsson \And Piersilvio De Bartolomeis \And  Issa J. Dahabreh \And Jesse H. Krijthe}

\aistatsaddress{TU Delft \And  ETH Zürich \And Harvard University \And TU Delft} ]

\begin{abstract}
  Randomized trials are typically designed to detect average treatment effects but often lack the statistical power to uncover individual-level treatment effect heterogeneity, limiting their value for personalized decision-making. To address this, we propose the QR-learner, a model-agnostic learner that estimates conditional average treatment effects (CATE) within the trial population by leveraging external data from other trials or observational studies. The proposed method is robust: it can reduce the mean squared error relative to a trial-only CATE learner, and is guaranteed to recover the true CATE even when the external data are not aligned with the trial. Moreover, we introduce a procedure that combines the QR-learner with a trial-only CATE learner and show that it asymptotically matches or exceeds both component learners in terms of mean squared error. We examine the performance of our approach in simulation studies and apply the methods to a real-world dataset, demonstrating improvements in both CATE estimation and statistical power for detecting heterogeneous effects.
\end{abstract}

\section{INTRODUCTION} \label{sec:introduction}

By randomly assigning the interventions of interest, randomized trials are unique in their ability to estimate causal effects with little reliance on untestable assumptions. This strength has made trials the preferred approach for evaluating interventions across many scientific domains. However, their high cost often limits sample size, which in turn limits the precision of statistical inferences that can be drawn from trial data. The problem is especially acute when the aim is not only to estimate an average treatment effect but also to characterize treatment effect heterogeneity~\citep{lagakos2006challenge}, a key step toward personalized decision-making for the population represented by the trial. A central quantity for this purpose is the conditional average treatment effect (CATE), which captures how treatment effects depend on individual-level covariates~\citep{kunzel2019metalearners}. However, the estimation of CATEs for different subgroups is more challenging than the estimation of average effects; therefore, the data from trials powered to detect average treatment effects are typically not adequate for the precise estimation of CATEs ~\citep{dahabreh2016using}. As a result, accurately estimating CATEs within a trial population remains a difficult yet important~challenge.

In recent years, there has been growing interest in augmenting trials with external data, mainly in the context of improving average treatment effect estimation~\citep{van2018including, jahanshahi2021use}. A key challenge in this setting is to properly account for differences between the trial population and the population underlying the external data~\citep{ung2024combining}. These differences raise a fundamental concern: whether causal quantities such as the CATE remain stable across the two populations -- a property known as transportability~\citep{bareinboim2016causal,dahabreh2019extending}. In this paper, we investigate the analogous problem of using external data, such as from another trial or an observational study, to augment the estimation of CATEs in the trial population, in settings where the populations underlying the trial and external data may be misaligned. Here, we focus on the setting where transportability does not necessarily hold and the external data may be subject to unmeasured confounding. Our objective is to leverage external data for CATE estimation while ensuring that using the external data does not harm the estimation compared to if we had used trial data alone in cases of misalignment between the underlying populations.

\paragraph{Contributions} We propose the QR-learner, a model-agnostic learner that improves estimation of the CATE in the population underlying a trial by leveraging external data from other trials or observational studies. We prove this learner is robust even when the external data are not aligned with the trial data: it recovers the true CATE even when external data come from a population which is not transportable with the trial population or are affected by uncontrolled confounding, while at the same time it can reduce the estimated CATE mean squared error compared to using trial data alone when the external data are sufficiently aligned (Sections~\ref{sec:class_pseudo_outcome}-\ref{sec:algorithm}). As another safeguard against potential harm from misaligned external data, we propose a procedure that combines the QR-learner with a trial-only CATE learner and prove that the combined learner asymptotically achieves a mean squared error that is no worse and potentially better than its component learners (Section~\ref{sec:combined}). Using simulations and real-world data from the Student/Teacher Achievement Ratio (STAR) project~\citep{word1990state,krueger1999experimental} we find that our method is robust when integrating external data that are not aligned with the trial data, and that it improves both CATE estimation mean squared error and the statistical power to detect treatment effect heterogeneity (Section~\ref{sec:experiment} and~\ref{sec:star}).

\section{RELATED WORKS}

Our work builds upon a rich and growing literature on CATE learners. Examples of approaches include adaptations of decision trees~\citep{athey2016recursive}, random forests~\citep{wager2018estimation}, and neural networks~\citep{shalit2017estimating}. Our proposed learner is most closely related to model-agnostic ``meta-learners''  which allow for estimating the CATE and nuisance models using any supervised learning algorithm~\citep{kunzel2019metalearners, nie2021quasi, kennedy2023towards}. However, most existing model-agnostic learners are tailored to settings where data are drawn from a single source, such as a single randomized trial or an observational~study.

More recently, several CATE learners have been proposed for multi-source settings, often relying on the assumption that the CATE is transportable across the underlying populations~\citep{hatt2022combining,wu2022integrative,shyr2023multi,wu2023transfer}. For example, \citet{schweisthal2024meta} propose a learner that constructs bounds on the transportable CATE under unmeasured confounding across different populations, while \citet{kallus2018removing} rely on the transportability assumption by estimating the CATE from a large, potentially confounded observational dataset and then apply a linear bias correction using data from a small randomized trial. Although some of these approaches might be used to estimate the trial-specific CATE under non-transportability, to our knowledge only \citet{asiaee2023leveraging,asiaee2023improving} explicitly address this setting; we discuss these methods in more detail later. \citet{yang2023elastic} studied the related problem of using multi-source data to perform valid statistical inference when estimating treatment effect heterogeneity under violations of transportability. However, their approach is restricted to a parametric linear working model for the CATE, whereas we focus specifically on optimizing the predictive performance (i.e., minimizing mean squared error) for a model-agnostic CATE learner which also allows for more flexible and nonparametric working models.

Finally, our work draws on recent developments in the trial augmentation literature for average treatment effect estimation in trials using data from an external population. These developments have emphasized robustness to integrating external data misaligned with the trial data~\citep{schuler2022increasing, huang2023leveraging,liao2023prognostic, de2025efficient,karlsson2024robust}. In particular, we will adapt ideas from the randomization-aware estimator framework proposed by \citet{karlsson2024robust} to construct  CATE learners that are robust to misaligned external data.

\section{PROBLEM SETTING} \label{sec:problem_setting}

\paragraph{Notation} Let $X \in \mathcal{X}$ denote baseline (pre-treatment) covariates; $S$ the binary indicator of data source ($S = 1$ for trial participants; $S = 0$ for individuals in the external data); $A$ the binary indicator for treatment assignment ($A = 1$ for the experimental treatment; $A = 0$ denotes the control); and $Y \in \mathcal{Y}$ the outcome (continuous, binary, or count). Throughout, we use italic capital letters to denote random variables and lowercase letters for their specific values. We write $f(\cdot)$ to denote the density functions of random variables.

\paragraph{Study design and data structure} We consider a non-nested trial design where the trial and external data are separately obtained and modeled as simple random samples from different populations, obtained with unknown and possibly unequal sampling fractions \citep{dahabreh2021study}. For observation $i$ with $S_i = s$, the data are modeled as i.i.d., conditional on study source, with the random tuple $O_i = (X_i, S_i = s, A_i, Y_i)$ for $i = 1, \ldots, n_s$, where $n_s$ is the number of observations from source $S = s$. The composite dataset has total sample size $n = n_1 + n_0$, where the proportions of trial and external participants in the composite dataset may not reflect the size of their underlying populations.  In the trial, treatment is randomly assigned according to the propensity score $e(X) = \Pr(A = 1 \mid X, S = 1)$ which is assumed to be known~\citep{rosenbaum1983central}. As $n \to \infty$, we assume the ratios of the trial and external data sample sizes to the total sample size converge to some constants, i.e., $n_s/n \to q_s \in(0,1)$.

\subsection{Identification of Causal Effects} To define the causal quantity of interest, we use potential outcomes~\citep{rubin1974estimating}. For individual $i$ and for $a\in\{0,1\}$, the potential outcome $Y_i^a$ denotes the outcome under intervention to set treatment $A$ to $a$, possibly contrary to fact. Our goal is to estimate the CATE in the population underlying the trial, 
\begin{equation} \label{eq:cate}
    \tau(x) = \E[Y^1 - Y^0 \mid X=x, S = 1]~.
\end{equation}
Under standard conditions, the CATE $\tau(x)$ is identifiable from data in the trial.
\begin{thmcond}[Consistency]\label{asmp:consistency}
If $A_i = a$, then $Y^a_i = Y_i$ for every individual $i$ and treatment $a \in \{0,1\}$.
\end{thmcond}
\begin{thmcond}[Strong ignorability in the trial population]\label{asmp:strong_ignorability_trial}
\textit{Positivity in trial:} for each treatment $a \in \{0,1\}$, if $f(x, S=1) \neq 0$, then $\Pr(A = a | X = x, S = 1) > 0$.
\textit{Conditional exchangeability in trial:} for each $a \in \{0,1\}$, $Y^a \indep A | (X, S=1)$. 
\end{thmcond}
Conditions~\ref{asmp:consistency} and~\ref{asmp:strong_ignorability_trial} are typically supported by a well-designed randomized trial and together suffice to identify the CATE as $\tau(x) = g_1(x) - g_0(x)$, where $g_a(x) = \E[Y \mid X = x, A = a, S = 1]$. However, it is common to assume additional conditions to enable identification and estimation of $\tau(x)$ using both the trial and external data.
\begin{thmcond}[Strong ignorability in the external population]\label{asmp:strong_ignorability_external}
    \textit{Positivity in external population:} for each treatment $a \in \{0,1\}$, if $f(x, S=0) \neq 0$, then $\Pr(A = a | X = x, S = 0) > 0$. \textit{Conditional exchangeability in external population:} for each $a \in \{0,1\}$, $Y^a \indep A | (X, S=0)$.
\end{thmcond}
\begin{thmcond}[Transportability]\label{asmp:transportability}
    For each $a \in \mathcal \{0,1\}$, $Y^a \indep S \mid X$.
\end{thmcond}
The above two conditions can be controversial, especially when the external data originate from an observational study,  because these conditions are uncertain and typically require substantial domain expertise to justify.
Notably, Conditions~\ref{asmp:consistency} to \ref{asmp:transportability} together have testable implications that can be empirically assessed to falsify them, see e.g.~\citet{hussain2023falsification,de2024detecting,dahabreh2024using}. This can be used in particular to evaluate  Conditions~\ref{asmp:strong_ignorability_external} and~\ref{asmp:transportability} because Condition~\ref{asmp:consistency} and~\ref{asmp:strong_ignorability_trial} are supported by the trial's experimental design. Nonetheless, performing such falsification tests remains an inherently difficult task~\citep{fawkes2025hardnessvalidatingobservationalstudies}.

\section{AUGMENTING TRIALS WITH EXTERNAL DATA}\label{sec:theory}

\subsection{A Class of Robust Pseudo-outcomes} \label{sec:class_pseudo_outcome}

Our goal is to learn a CATE function from a class of candidates $\mathcal{F}$ that minimizes the population risk relative to the true CATE function, namely $\argmin_{\tilde\tau\in\mathcal{
F}}R^*(\tilde{\tau})$ where $R^*(\tilde\tau) = \E[(\tau(X) - \tilde\tau(X))^2 \mid S = 1]$.
However, as we cannot minimize $R^*(\tilde\tau)$ directly because the true CATE $\tau(X)$ is unknown, we study the class of CATE learners obtained by minimizing a pseudo-risk~\citep{foster2023orthogonal},
\begin{equation}
    \argmin_{\tilde\tau\in\mathcal{F}} R(\tilde\tau; \eta)\label{eq:pseudo_risk_obj}
\end{equation}
where $R(\tilde\tau; \eta) = \E[(\psi(O;\eta) - \tilde\tau(X))^2 \mid S = 1]$.
Here, we introduce an auxiliary random variable, sometimes referred to as a pseudo-outcome: 
\begin{equation}\label{eq:pseudo_outcome}
\begin{split}
\psi(O_i; \eta) &= 
    \frac{A_i - e(X_i)}{e(X_i)(1-e(X_i))}
        \bigl(Y_i - h_{A_i}(X_i)\bigr) \\
    &\quad + h_1(X_i) - h_0(X_i)
\end{split}
\end{equation}
which is indexed by some nuisance models $\eta = \{h_1, h_0\}$, where $h_1 : \mathcal{X} \rightarrow \mathbb{R}$ and $h_0 : \mathcal{X} \rightarrow \mathbb{R}$ are real-valued functions defined on the covariate space $\mathcal{X}$. When clear from context, we omit the arguments and write $\psi_i = \psi(O_i; \eta)$ and $\widehat{\psi}_i = \psi(O_i; \hat\eta)$ to denote the pseudo-outcomes computed using either the nuisance models $\eta$ or replacing the nuisance models with their estimates~$\hat\eta$.

Depending on our choice of $\eta$, we obtain different CATE learners when solving~\eqref{eq:pseudo_risk_obj}. For instance, if $\eta = \{0,0\}$, we obtain the (inverse) propensity weighted learner, refereed to as the PW-learner by~\citet{curth2021nonparametric}. Meanwhile, if $\eta=\{g_1, g_0\}$, where $g_a=\E[Y\mid X, A=a, S=1]$, we obtain the DR-learner~\citep{kennedy2023towards}. More generally, for any choice of $\eta$, we can prove an important robustness property of $R(\tilde\tau;\eta)$ guaranteed by the trial's randomized design.
\begin{thmthm}\label{thm:robustness_pseudo_outcome}
    Under Conditions~\ref{asmp:consistency} and~\ref{asmp:strong_ignorability_trial} where the propensity score $e(X)$ is known, for any fixed specification of the nuisance models $\eta_{\text{fixed}}$, the minimization problem in~\eqref{eq:pseudo_risk_obj} always yields the true CATE as its unique solution provided that $\tau\in\mathcal{F}$; that is, $\tau = \argmin_{\tilde\tau \in \mathcal{F}} R(\tilde\tau; \eta_{\text{fixed}})$. 
\end{thmthm} 
Although this result has appeared in the literature before (see, for example, \citet{morzywolek2023weighted} and references therein), we provide a derivation in Appendix~\ref{app:robustness_proof} for completeness. We denote the nuisance models $\eta_{\text{fixed}}$ with a fixed specification to emphasize that they must be chosen independently of the dataset used to compute the pseudo-outcome; this requirement can be satisfied using cross-fitting, which we describe in the following~subsections.

Recognizing the central role of using the known propensity score in the above theorem, we refer to the pseudo-outcomes in~\eqref{eq:pseudo_outcome} constructed using the known propensity score as \textit{randomization-aware} pseudo-outcomes~\citep{karlsson2024robust}. This framing allows us to consider a broader class of pseudo-outcomes for robust CATE estimation, distinguished by varying specifications of the nuisance models $\eta$. While Theorem~\ref{thm:robustness_pseudo_outcome} guarantees that $R(\tilde\tau; \eta)$ is a proper model selection criterion for the CATE regardless of the choice of $\eta$, as we will discuss next, the choice of $\eta$ ultimately still plays a crucial role in estimating the CATE based on the observed data. This happens as we transition to the sample analog version of the minimization problem in~\eqref{eq:pseudo_risk_obj}.

\begin{thmrem}
    While our main focus is on the pseudo-risk and pseudo-outcomes in~\eqref{eq:pseudo_risk_obj} and~\eqref{eq:pseudo_outcome}, another relevant pseudo-risk that may share similar robustness properties when having access to the true propensity score $e(X)$ is the one used by the R-learner~\citep{nie2021quasi}. In Appendix~\ref{app:R-learner}, we outline potential connections between our results and the R-learner pseudo-risk, particularly in the special case where the propensity score $e(X)$ is constant.
\end{thmrem}

\subsection{Using External Data to Improve CATE Model Selection in the Trial Population}

To estimate the CATE from observed data, we must consider the sample analog of $R(\tilde\tau; \eta)$, defined as
$
    \widehat{R}(\tilde\tau; \hat\eta) =\frac{1}{n_1}\sum_{i: S_i=1} \left( \psi\left(O_i; \hat\eta\right) - \tilde\tau\left(X_i\right) \right)^2.
$
We then obtain the CATE estimate by solving $\hat\tau = \argmin_{\tilde\tau\in\mathcal{F}} \widehat{R}(\tilde{\tau}; \hat\eta)$.

Although the robustness property of randomization-aware pseudo-outcomes discussed earlier might suggest that the choice of nuisance models $\eta$ is inconsequential, we will show that this is not the case. Because the model selection criterion $\widehat{R}(\tilde\tau;\hat\eta)$ is a sample average, in finite samples, this criterion can choose suboptimal CATE models and, importantly, this behavior is influenced by the choice of nuisance models $\eta$. To understand this, we first note that we can decompose the sample analog pseudo-risk as
\begin{equation*}
    \begin{split}
         \widehat{R}(\tilde\tau;\hat\eta)= \frac{1}{n_1}\sum_{i:S_i=1} \Big[ \,
           & (\tau(X_i)-\tilde\tau(X_i))^2 \\
         - & 2(\tau(X_i)-\tilde\tau(X_i))(\widehat{\psi}_i-\tau(X_i)) \\
         + & (\widehat{\psi}_i-\tau(X_i))^2
        \Big]~.
    \end{split}
\end{equation*}
From the above decomposition, which we derive in Appendix~\ref{app:decomposition}, we see that $\widehat{R}(\tilde\tau;\hat\eta)$ consists of three parts: a first term that in expectation equals the population risk $R^*(\tilde\tau) = E[(\tau(X) - \tilde\tau(X))^2 \mid S = 1]$; a second term that introduces model selection uncertainty; and a third term which is independent of the candidate model $\tilde\tau$. Therefore,  we may still end up selecting a suboptimal $\tilde\tau$ due to the second term influencing $\widehat{R}(\tilde\tau;\hat\eta)$. 

To see why the second term is problematic for model selection, consider comparing two candidate models $\tilde\tau_1$ and $\tilde\tau_2$. If we observe $\widehat\Delta_{12}=\widehat{R}(\tilde\tau_1;\hat\eta)-\widehat{R}(\tilde\tau_2;\hat\eta)>0$, we would conclude that $\tilde\tau_2$ is better than $\tilde\tau_1$; we would make the opposite decision when $\widehat\Delta_{12}<0$, or remain inconclusive when $\widehat\Delta_{12}=0$. However, note that only the third term from the decomposition cancels out in the risk difference $\widehat\Delta_{12}$. This means that in addition to the first term, which captures the true error relative to the true CATE, the second term can also influence our decision about which candidate model performs better.

To improve $\widehat{R}(\tilde\tau;\hat\eta)$ as a model selection criterion, a natural strategy is to choose $\eta$ to minimize the variance of the problematic second term. The next result provides insight into how this can be achieved (see proof in Appendix~\ref{app:variance_bound}).

\begin{thmlem} \label{lem:variance_bound}
Under Conditions~\ref{asmp:consistency} and~\ref{asmp:strong_ignorability_trial}, assume the propensity score $e(X)$ is known and the nuisance functions $\hat\eta=\{\hat{h}_1,\hat{h}_0\}$ are estimated on a dataset independent of that used to compute the pseudo-outcomes $\widehat{\psi}$. Define $\epsilon := (\tau(X)-\tilde\tau(X))(\widehat{\psi}-\tau(X))$, then we have that $\E[\epsilon\mid S=1]=0$ and $\V(\epsilon \mid S=1)\;\leq\;\tilde{C}\Bigl[2\{L_1(\hat{h}_1)+L_0(\hat{h}_0)\}+\tilde\sigma^2\Bigr]$
where
\begin{align*} \label{eq:mse_f_a}
    L_a(\hat{h}_a) &=  \E\left[w_a(X) \left(Y-\hat{h}_a\left(X\right)\right)^2 \mid A=a, S=1\right] \\
    \tilde\sigma^2 & = \V(Y^1-Y^0\mid S=1)-\V(\tau(X)\mid S=1) \\
    \tilde{C} & =\max_{x\in\mathcal{X}}(\tau(x)-\tilde\tau(x))^2
\end{align*}
with $w_a(X)=\left(\frac{1-e(X)}{e(X)}\right)^{2a-1}$ for $a\in\{0,1\}$.
\end{thmlem}

The above result shows that a weighted mean squared error of the nuisance models $\hat\eta = \{\hat{h}_1, \hat{h}_0\}$ appears in an upper bound on the variance of the terms responsible for model selection uncertainty when using $\widehat{R}(\tilde\tau; \hat\eta)$. This motivates our approach for selecting $\eta$: choose it to directly minimize both $L_1$ and $L_0$.
Similar approaches to selecting $\eta$ have appeared in prior work in single-source settings. \citet{cao2009improving} addressed a related problem of mean estimation with missing data, while \citet{saito2020counterfactual} considered conditional average treatment effect estimation.

The tightness of the bound in the above lemma depends on the candidate model $\tilde\tau$ for the CATE. Specifically, it is proportional to the non-negative constant $\tilde{C} = \max_{x \in \mathcal{X}} (\tau(x) - \tilde\tau(x))^2$. Therefore, when the class of candidate models $\mathcal{F}$ contains good approximations of the true CATE $\tau$ and $\tau$ itself is bounded, $\tilde{C}$ is expected to be small, resulting in a relatively tight upper bound.

In the next result, we show how external data can be used to minimize the identified upper-bound in Lemma~\ref{lem:variance_bound}. To achieve this, we invoke the assumptions of strong ignorability in the external data (Condition~\ref{asmp:strong_ignorability_external}) and transportability (Condition~\ref{asmp:transportability}). However, because these conditions are uncertain and may not hold in many settings, we recall that these conditions are not needed to ensure the robustness properties of the randomization-aware pseudo-outcomes. We later discuss a strategy for further improving robustness against these violations.
\begin{thmthm}\label{thm:variance_rewritten} 
Under Conditions~\ref{asmp:strong_ignorability_external} and~\ref{asmp:transportability}, in addition to those of Lemma~\ref{lem:variance_bound}, we can express
\begin{equation}\label{eq:f_a_rewritten}
    L_a(\hat{h}_a) = \E\Bigl[\tilde{w}_a(X)\,(Y-\hat{h}_a(X))^2 \mid A=a\Bigr]
\end{equation}
where $\tilde{w}_a(X) = \frac{\Pr(S=1 \mid X, A=a)}{\Pr(S=1 \mid A=a)} \left(\frac{1-e(X)}{e(X)}\right)^{2a-1}$.

\end{thmthm}
The above theorem, proven in Appendix~\ref{app:variance_rewritten}, shows that the function $L_a(\hat{h}_a)$, which appeared in the upper-bound of the variance $\V(\epsilon \mid S=1)$, can be rewritten in terms of all observed data from both the trial and external populations. Unlike the expression for $L_a(\hat{h}_a)$ in Lemma~\ref{lem:variance_bound}, we no longer condition on $S$. With this result, we can proceed with proposing a novel CATE learner that chooses the nuisance models $\eta$ to minimize this upper-bound with the help of external data. 

\begin{thmrem}
The results in Theorem~\ref{thm:variance_rewritten} are reminiscent of results from the transfer learning literature, where one express the expected mean squared error of a prediction function for data within some domain by reweighting data from another domain~\citep{shimodaira2000improving,weiss2016survey}. To obtain this type of result in transfer learning, one typically assumes that the conditional distribution of labels remains constant across the two domains; this is analogous to how the transportability condition put certain requirements on the conditional distribution of the potential outcomes in our setting. However, unlike this literature, our goal is to be explicitly robust to violations of this condition, for which we propose a solution to this problem later in Section~\ref{sec:combined}.
\end{thmrem}

\subsection{The QR-learner Algorithm} \label{sec:algorithm}

In this section, we introduce a novel method for estimating the CATE, which we call the \textit{Quasi-optimized Randomization-aware} learner, or QR-learner. This method is model-agnostic, allowing it to use any supervised learning algorithm for estimating the CATE and nuisance models. It follows a two-stage procedure: In the first stage, it solves an optimization problem to select $\eta$ using both trial and external data that minimizes the identified upper bound related to the model selection uncertainty in finite samples. We call this step quasi-optimized because it targets an upper bound rather than the variance directly. In the second stage, the method regresses randomization-aware pseudo-outcomes on the covariates using only trial data to estimate the CATE. The first stage reduces finite-sample uncertainty for model selection, while the second stage leverages the robustness of the randomization-aware pseudo-outcomes to target the true CATE in the trial population.

To prevent overfitting and ensure that $\hat\eta$ is estimated independently of the data used for pseudo-outcomes, we employ a cross-fitting procedure that partitions the data into $\mathcal{D}^1 \cup \mathcal{D}^2$, stratified by treatment $A$ and study population $S$. In the first stage, we use both the trial and external data in $\mathcal{D}^1$ to estimate $\hat\eta^* 
= \{\hat{h}_1^*, \hat{h}_0^*\}$ where each component $\hat{h}_a^*$ is obtained by solving the optimization problem
\begin{equation}\label{eq:optimization_h}
\min_{h_a \in \mathcal{H}} \sum_{i \in \mathcal{D}_a^1} \widehat{\pi}_a(X_i) \Bigl(\frac{1-e(X_i)}{e(X_i)}\Bigr)^{2a-1} \bigl(Y_i - h_a(X_i)\bigr)^2
\end{equation}

where the sum is taken over $\mathcal{D}_a^1=\{i\in \mathcal{D}^1 : A_i=a\}$, $\mathcal{H}$ is the model class under consideration for the nuisance models $\eta=\{h_1,h_0\}$, and $\widehat{\pi}_a$ is an estimator from the model class $\mathcal{G}$ for the probability of trial participation $\Pr(S=1\mid X=x,A=a)$ which is also estimated using $\mathcal{D}_a^1$. Then, in the second stage, the pseudo-outcomes $\psi(O_i;\hat\eta^*)$ are computed on $\mathcal{D}^2$ using the estimated nuisance models $\hat\eta^*$. We estimate the CATE using $\mathcal{D}^2$ by solving $\hat\tau=\argmin_{\tilde\tau\in\mathcal{F}}\sum_{i\in \mathcal{D}^2: S_i=1}(\psi(O_i;\hat\eta^*) - \tilde\tau(X_i))^2$. To efficiently use all available data, we reverse the roles of the splits to obtain a second CATE estimator, and then take the average of the predictions from the two resulting estimators; this procedure naturally extends to more than two data splits if desired. Pseudo-code for the QR-learner is also provided in Appendix~\ref{app:pseudo_code}. 

We now discuss several noteworthy remarks about the QR-learner. First, we recommend using a linear logistic regression with a cross-validated ridge penalty for estimating $\widehat{\pi}_a$. The reason we use a linear model is that, although the external sample size is large, the trial sample size may be much smaller, so a more flexible model could still overfit or have poor calibration. It is also known that maximum likelihood estimation of logistic regression can be biased in small-sample settings or when the number of observations from one of the events is rare. Adding a shrinkage penalty, as we do here, can reduce this type of bias; see, for example,~\citet{firth1993bias} and~\citet{leitgob2020analysis} for more detailed discussion.

Second, while our primary focus is to predict the CATE well with respect to the risk $R^*(\hat\tau)=\E[(\tau(X)-\hat\tau(X))^2\mid S=1]$, we note that in some cases, it may also be possible to use the estimated function $\hat\tau(X)$ obtained from the QR-learner for inference about the CATE. Although a full theoretical treatment is beyond the scope of this work, we provide supporting arguments in Appendix~\ref{app:QR_learner_guarantees} to explain why this may be justified, and later show in a simulation study that this is feasible.

Third, our motivation for proposing the QR-learner stems from Theorem~\ref{thm:variance_rewritten}, which shows that minimizing the objective in~\eqref{eq:optimization_h} can improve CATE model selection. Here we provide another argument for why the QR-learner may outperform a trial-only learner, specifically the DR-learner. Both learners differ only in how their outcome nuisance models are estimated, and it is known that the DR-learner's convergence rate depends on errors in both estimating the CATE function in the second stage and the outcome nuisance models in the first stage~\citep{kennedy2023towards}. The QR-learner can improve upon the DR-learner by estimating these outcome nuisance models more accurately using both trial and external data when populations are well-aligned. With sufficiently large external data, outcome nuisance estimation errors become negligible, and we could expect the QR-learner to outperform the DR-learner, which is also what we observe in our simulations.

Finally, the CFACE learner proposed by~\citet{asiaee2023leveraging} shares close similarities with the QR-learner. Their method differs from ours in the first stage by estimating nuisance models $\eta = \{m^*, m^*\}$, where $m^*(x) = e(x) \cdot \mu_0(x) + (1 - e(x)) \cdot \mu_1(x)$ with $\mu_a(x) = \mathbb{E}[Y \mid X = x, A = a, S = 0]$. This formulation comes from solving $\argmin_m \V\left( \psi(O; \eta=\{m,m\}) \mid X=x, S=1\right)$, with nuisance models estimated entirely from external data. Their approach works well when external data are abundant and aligned with the trial population, but can fail if the populations differ substantially or external data are limited. This pitfall was noted by the same authors in~\citet{asiaee2023improving}, where they propose another learner called R-OSCAR. This learner differs more substantially from ours and instead resembles the bias correction approach of~\citet{kallus2018removing}, but tailored specifically for estimating the CATE in the underlying population of a randomized trial. We further discuss the procedures of both CFACE and R-OSCAR in Appendix~\ref{app:asiaee}.

\subsection{Combining CATE Learners}\label{sec:combined}

The success of the optimization in~\eqref{eq:optimization_h} relies on two conditions that are not necessary for the identification of the CATE in the trial population. First, it requires that the external data are aligned with the trial data: that is, Conditions~\ref{asmp:strong_ignorability_external} and \ref{asmp:transportability} hold. Second, for each $a\in\{0,1\}$, the estimator $\widehat\pi_a(x)$ needs to be a correctly specified model of the probability $\Pr(S = 1 \mid X=x, A = a)$. As a result, while the QR-learner always targets the correct CATE in the trial population, it may perform worse in finite samples, in terms of mean squared error, than a CATE learner based on the trial data alone. To address this issue, we therefore propose a variant which combines the potential benefits from using external data via the QR-learner with the additional robustness from a trial-only learner. Specifically, we use the DR-learner~\citep{kennedy2023towards} obtained by regressing randomization-aware pseudo-outcomes $\psi(O;\hat\eta=\{\hat{g}_1, \hat{g}_0\})$ on the covariates $X$ where $\hat{g}_a$ estimates $\E[Y\mid X,A=a, S=1]$  using only trial~data.

Our proposed combined learner is defined as
$$\hat\tau(x;\lambda) := \lambda\cdot \hat\tau_{\text{QR}}(x) + (1-\lambda)\cdot \hat\tau_{\text{DR}}(x), \; \lambda\in [0,1]~,$$
where $\hat\tau_{\text{QR}}$ is the estimator from the QR-learner using both trial and external data, and $\hat\tau_{\text{DR}}$ is the estimator from the DR-learner using trial data alone; this combined learner can be viewed as an instance of stacked regression~\citep{breiman1996stacked} for CATE estimation. We recommend using the DR-learner as the trial-only learner here because it is also robust to misspecification of the outcome models, even though the theoretical guarantees in this section also hold if we replace it with another CATE learner fitted using only the trial data.

Recall the population risk introduced in Section~4.1, $R^*\!\left(\tilde\tau\right)=\mathbb{E}\!\bigl[(\tau(X)-\tilde\tau(X))^{2}\mid S=1\bigr].$ 
For the combined learner we write $R^*(\lambda):=R^*\!\bigl(\hat\tau(\,\cdot\,;\lambda)\bigr)$ with $R^*_{\mathrm{DR}}=R^*(0)$ and $R^*_{\mathrm{QR}}=R^*(1)$.
Our goal in this section is to choose $\lambda$ such that $R^*(\lambda)$ is no larger than
$\min\{R^*_{\mathrm{DR}},R^*_{\mathrm{QR}}\}$.
 We first show that for the oracle weights the mean squared error of the combined estimator is no worse than the individual components (see proof in Appendix~\ref{app:oracle_weight_combined}).
\begin{thmlem}\label{lem:oracle_weight_combined}
Let $\lambda^{\star}=\arg\min_{\lambda \in \Lambda} R^*(\lambda)$ be the oracle weight then $R^*(\lambda^{\star}) \le \min\{R^*_{QR}, R^*_{DR}\}$.
Furthermore, if $\E[(\hattauQR(X)-\hattauDR(X))^{2}\mid S=1] > 0$, then the oracle weight $\lambda^{\star}$ is unique. 
\end{thmlem}

In practice, the population risk cannot be computed directly, so we minimize the sample analog pseudo-risk $\widehat R (\lambda) = \frac{1}{n_1} \sum_{i:S_i=1} (\widehat\psi_i-\tauHat(X_i;\lambda))^2.$ We show that an empirical estimate of the oracle weight $\lambda^{\star}$ can be obtained via cross-validation and achieves the same risk asymptotically as the oracle. Let the trial data be partitioned into $K$ mutually exclusive folds. For each observation $i$, let $\hat\tau_{\text{DR}}^{(-i)}$ and $\hat\tau_{\text{QR}}^{(-i)}$ denote the learners trained on data excluding the fold containing $i$, and define $\hat\tau^{(-i)}(x;\lambda):=\lambda\cdot \hattauQR^{(-i)} + (1-\lambda)\cdot \hattauDR^{(-i)}$. We consider computing the pseudo-outcomes with $\eta=\{0,0\}$, yielding a model-free combination approach.
The $K$-fold cross-validated pseudo-risk estimate is $\widehat R_{CV}(\lambda) = \frac{1}{n_1} \sum_{i=1}^{n_1}\bigl( \psi_i - \hat\tau^{(-i)}(x;\lambda) \bigr)^2$. The cross-validated weight is then obtained as $\hat\lambda^\star=\arg\min_{\lambda\in \Lambda}\widehat R_{CV}(\lambda)$.
\begin{thmthm} \label{thm:cv_risk_combined}
   Assume that the pseudo‑outcome $\psi(O;\eta=\{0,0\})$ and the base learners $\hattauQR(X)$ and $\hattauDR(X)$ are uniformly bounded. Then, 
   we have $R^*(\hat\lambda^{\star})\;\le\;R^*(\lambda^\star)\;+\;o_p(1)
     \;\le\;\min\{R^*_{DR}, R^*_{QR}\}\;+\;o_p(1)$
   where $o_p(1)$ converges to $0$ as $n\rightarrow\infty$.
\end{thmthm}

The above result describes the oracle excess risk of the combined learner and guarantees that selecting the parameter $\lambda$ via cross-validation yields a combined learner whose risk differs from the best possible risk by a vanishing term. In other words, the cross-validated choice is asymptotically as good as the oracle-optimal choice and, in large samples, performs at least as well as the better of the two candidate learners. 
Note that our setting does not require the two component learners to converge to the same CATE function. If they use different model classes and converge to different functions, a linear combination may be strictly better than either component learner. In the special case where both learners converge to the same function such that $R^*_{DR}=R^*_{QR}$, the combined learner also converges to this function.

As a final remark, we note that an alternative approach to combine the QR- and DR-learner would be to take the pseudo-outcomes from their respective first stages and fit a CATE model to a linear combination of these pseudo-outcomes. However, this is a less general solution since our formulation allows the theoretical guarantees in this section to hold for any trial-only learner, including those that do not rely on learning the CATE using pseudo-outcomes such as the S- or T-learner~\citep{kunzel2019metalearners}.

\section{SIMULATION STUDY} \label{sec:experiment}
We conduct a series of simulations to evaluate the performance of our proposed method against several baselines, focusing on CATE prediction mean squared error. We also assess statistical power to detect treatment effect heterogeneity through interaction tests between the treatment and a hypothesized effect modifier, an analysis commonly performed in real-world trials. For this second evaluation, we consider only methods that support statistical inference.
The code to reproduce our experiments is available in the public Github repository \url{https://github.com/RickardKarl/robust-trial-CATE-augmentation}.

\paragraph{Baselines} In addition to our proposed QR-learner and combined learner, we apply the DR-learner with known propensity scores using only trial data~\citep{kennedy2023towards}; the T-learner, which computes CATE estimates as $\hat{g}_1(x) - \hat{g}_0(x)$ with the same nuisance models that are used in the DR-learner; a pooled variant of the T-learner obtained by estimating $\E[Y\mid X,A=1]-\E[Y\mid X,A=0]$ using both the trial and external data; the method proposed by~\citet{asiaee2023improving} called R-OSCAR, in addition to a method from a previous version of their manuscript called CFACE (see~\citet{asiaee2023leveraging}; an extended discussion on their differences is found in Appendix~\ref{app:asiaee}); and the linear additive bias correction method of~\citet{kallus2018removing} which we refer to as KSP (after the authors' initials). Implementation details are provided in Appendix~\ref{app:implementation_details}.

\subsection{Influence of External Data Sample Size and Population Misalignment} \label{sec:simulation_rmse} 
First, we assess the root mean squared error (RMSE) of CATE predictions in a setting with a fixed trial size while varying the size of the external dataset. We consider two scenarios: (i) an idealized setting in which both Conditions~\ref{asmp:strong_ignorability_external} and~\ref{asmp:transportability} hold -- i.e., the external data are unconfounded and transportability holds -- and (ii) a more realistic, challenging setting where these assumptions are violated. We use gradient boosting regressors to estimate the outcome nuisance components and a linear model to estimate both the trial participation probability and the CATE; this modeling choice aligns with the data-generating process, where the baseline outcome is a highly nonlinear function of the covariates, while the CATE is linear.

Table~\ref{tab:n1_250_rmse} shows that our proposed methods consistently achieve the lowest or near-lowest RMSE across all settings, with performance improving as the external sample size increases. The trial-only DR- and T-learners performs no better than simply predicting an estimated average treatment effect (ATE) for all individuals, indicating the difficulty of this task based on trial data alone. While all integrative methods perform best when Conditions~\ref*{asmp:strong_ignorability_external} and~\ref*{asmp:transportability} hold, it is noteworthy when these conditions are violated that our methods exhibit RMSEs comparable to that of the trial-only DR-learner. CFACE shows a similar pattern, except when the external dataset is small $(n_0=100)$, where it performs worse than both the trial-only DR-learner and the ATE predictor. This suggests that their nuisance model fitting is somewhat less robust than ours, as it can underperform relative to the best trial-only CATE learner. Interestingly, R-OSCAR and KSP underperform in all settings; both of these methods rely on fitting separate functions to model the bias between the trial and external population. The difficulty of this bias modeling when the trial size is small could likely explain why both these methods fail.

When examining average prediction bias (full results in Appendix~\ref{app:additional_experiments}), we find that all methods are largely unbiased except the pooled T-learner. This is expected since we fit linear CATE models that match the true CATE function. Thus, the observed RMSE gains among the integrative methods using external data primarily stem from variance reduction rather than bias reduction, provided the model is correctly specified.

\begin{table*}[t]
    \centering 
    \caption{\small Average root mean squared error reported over 500 repeated runs from the simulation study with a trial sample size $n_1=250$ under different scenarios. Lowest number for each scenario is marked with bold.}
    \label{tab:n1_250_rmse}
    \resizebox{0.98\linewidth}{!}{%
    \input{tables/n1_250_rmse}
    }
\end{table*}

\subsection{Assessing  Statistical Power to Detect Interaction Effects} \label{sec:simulation_power} 
We next assess statistical power to detect treatment effect heterogeneity via interaction tests between the treatment and a hypothesized effect modifier when integrating external data while transportability (Condition~\ref{asmp:transportability}) is violated. This analysis is conducted in a setting with five observed covariates $X$, where one is a potential effect modifier $Z\subseteq X$. We consider two scenarios: one in which the effect modifier is present, and another where it is absent. Methods based on the T-learner, R-OSCAR, KSP and the combined learner are excluded, as they do not support inference on the estimated CATE model $\hat{\tau}$. For the remaining methods, we regress $\hat{\tau}$ on $Z$ and perform a two-sided test on the $Z$ coefficient. As a baseline, we also include a simpler test aligned with standard practice in trial analyses, fitting a linear regression of $Y$ on $(A, Z, A \cdot Z)$, testing the $A \cdot Z$ interaction using either only the trial data or the pooled data. We use a significance level $\alpha=0.05$. Full implementation details of the tests are provided in Appendix~\ref{app:statistical_tests}.  

\begin{figure*}[t]
    \centering
    \begin{subfigure}[b]{0.51\linewidth}
        \centering
        \includegraphics[width=\linewidth]{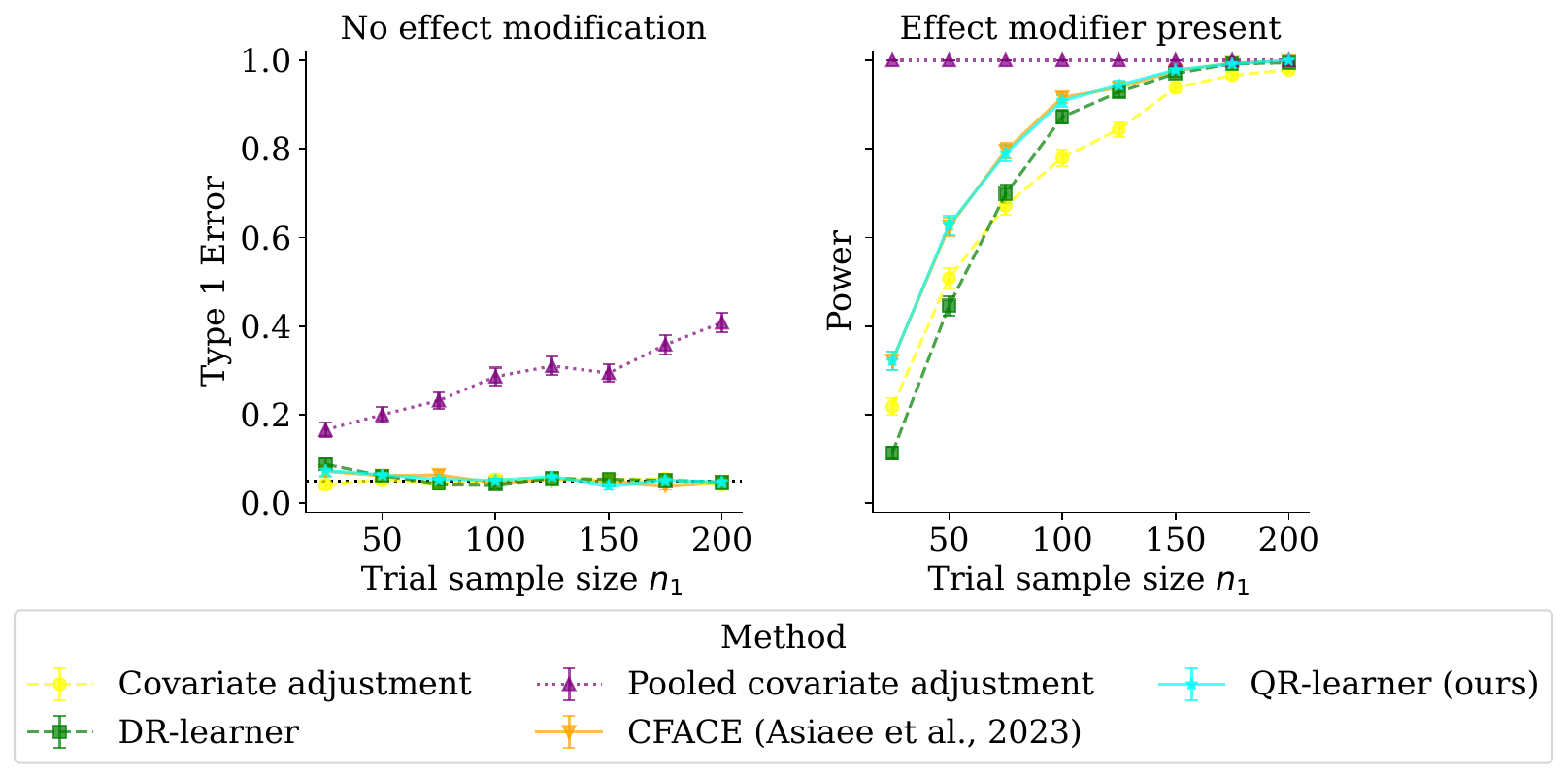}
        \caption{}
        \label{fig:power_comparison_effect}
    \end{subfigure}
    ~
    \begin{subfigure}[b]{0.4\linewidth}
        \centering    \includegraphics[width=\linewidth]{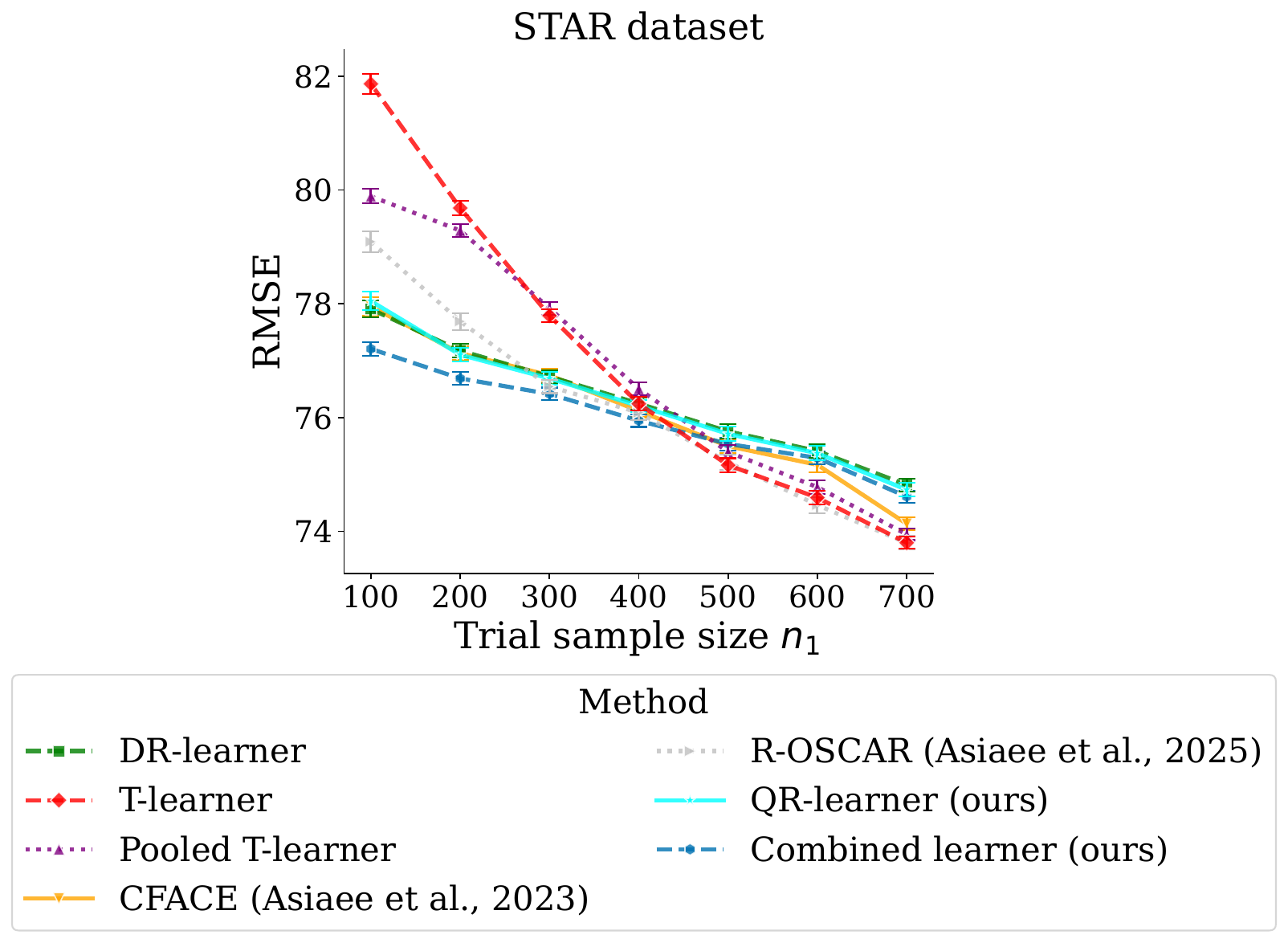}
        \caption{}
        \label{fig:star_urban_target}
    \end{subfigure}
    \caption{\small \textbf{(a)}: Evaluating type 1 error (lower better) and  power (higher better) in the simulation study with the methods applicable for statistically testing for the presence of an effect modifier as sample size in trial increases, reported over 500 repeated runs. \textbf{(b)} We evaluate the RMSE on the STAR dataset when increasing the trial sample size with a fixed external sample size of $n_0=1000$. We report the average RMSE and standard error over 200 repeated runs.
    }
    \label{fig:combined}
\end{figure*}

The results in Figure~\ref{fig:power_comparison_effect} show that when varying the trial sample size with the external dataset size fixed at $n_0 = 1000$, all methods maintain nominal type I error in the absence of an effect modifier except for the pooled covariate adjustment which can be explained by transportability being violated. Among the methods with nominal type 1 error, the QR-learner and CFACE consistently improve the power by about 10 to 20 percentage points to detect the effect modifier when it is present.

\section{CASE STUDY: STAR DATASET}\label{sec:star}
We use data from the Tennessee Student/Teacher Achievement Ratio (STAR) project~\citep{word1990state,krueger1999experimental}, a large-scale randomized trial on the effects of class size on student performance. The dataset is divided into two populations based on school location: rural and urban schools. Since both originate from a randomized study, strong ignorability is expected to hold within each population. However, outcome distributions differ across rural and urban schools, and by deliberately omitting school location from the observed covariates, we create a setting where transportability between the two populations is violated. Additional details on the data are provided in Appendix~\ref{app:star}.
Performance is measured using RMSE on a held-out test set from the target population, where the pseudo-outcome $\psi(O_i; \eta = \{0, 0\})$ serves as a proxy for the true CATE. Although this proxy has high variance, the use of the known trial propensity score ensures that the estimate aligns in expectation with the true population risk, as discussed in Section~\ref{sec:theory}. Gradient boosting is used for estimating the outcome nuisance models, and ridge-penalized (logistic) linear models to estimate both the trial participation probability and the CATE.

We evaluate the RMSE of each method as the trial sample size increases, while fixing the external data size at $n_0=1000$. In Figure~\ref{fig:star_urban_target}, urban schools are treated as the target trial population ($S=1$) and rural schools as the external population ($S=0$); the flipped setting is reported in Appendix~\ref{app:additional_experiments}. KSP performed significantly worse than the other methods and is therefore omitted from the main figure for ease of visualization, though it is included in the appendix. The integrative methods mostly benefit from external data when the trial sample is small, with gains diminishing as the trial grows. The combined learner consistently outperforms its component learners, the DR- and QR-learners, as predicted by our theory, and also outperforms CFACE and R-OSCAR at smaller sample sizes. The QR-learner offers no improvement over the DR-learner in this setting, though in the flipped setting we observe that it improves over the DR-learner.
When the trial size becomes large, the trial-only and pooled T-learners, along with CFACE and R-OSCAR, achieve the lowest RMSE. The margin between these approaches and the DR-learner also grows, although this margin depends on whether the target population is urban or rural. This suggests that the advantage may reflect asymmetries between the two populations rather than a general property of the methods. Overall, the results show that our proposed CATE learners can effectively leverage external data to improve prediction accuracy.

\section{DISCUSSION} \label{sec:discussion}

Our experimental findings demonstrate that the proposed learners for estimating the CATE using external data can effectively reduce the mean squared error of CATE estimates, while remaining robust in scenarios where the external data has unmeasured confounders or transportability is violated. Notably, in cases where the DR-learner failed to outperform a simple baseline that predicts the average treatment effect -- thus providing limited value -- our proposed learners were still able to achieve better accuracy. This highlights the potential value of incorporating external data into analyses of heterogeneous treatment effects in randomized trials.

While our proposed methodology performs as intended, it is important to acknowledge its limitations, particularly given its potential impact on real-world decision-making in public health and policy. First, we observe an upper limit to the incremental gain of incorporating additional external data. This is because the external data primarily improves the first-stage estimation of the QR-learner, whereas the second stage relies solely on the randomized trial and is therefore constrained by the trial sample size.  Second, although we observe that minimizing the upper bound in Lemma~\ref{lem:variance_bound} improves CATE accuracy empirically and our method appears robust, it remains of interest to explore under what scenarios tighter bounds can be obtained or more direct strategies for improving CATE estimation can be~developed. 

% Next, our theoretical results apply to any outcome type, whether continuous, binary, or count. However, the resemblance of~\eqref{eq:optimization_h} to a weighted squared error loss suggests that the QR-learner's objective is most naturally suited for continuous outcomes. For binary outcomes, one can ignore this structure and still minimize the same squared loss. Alternatively, restricting $h_a$ to binary outputs transforms the objective in~\eqref{eq:optimization_h} into a weighted zero-one loss, but this may be difficult to minimize due to its non-differentiability. While both approaches could, in principle, minimize the upper bound in Lemma~\ref{lem:variance_bound} and improve CATE estimation, their practical impact remains poorly understood and warrants further investigation, particularly given the prevalence of non-continuous outcomes in real-world applications.

%Finally, although our primary focus was on minimizing the mean squared error of CATE predictions, we observed promising results in simulations using the QR-learner for statistical testing of interaction effects within the trial population. Since this approach aligns more closely with standard practices in clinical and experimental research, it would be valuable to further investigate the scenarios and specific conditions under which it is expected to perform well. In particular, exploring power and sample size calculations based on the available external data could provide practical guidance for applying this method in real-world studies.

\section*{Acknowledgments}
Research reported in this work was facilitated by the computational resources and support of the Delft AI Cluster (DAIC) at TU Delft. Furthermore, this work was supported by Patient-Centered Outcomes Research Institute (PCORI) award ME-2021C2-22365, National Library of Medicine (NLM) award R01LM013616, and National Heart, Lung, and Blood Institute (NHLBI) award R01HL136708. The content is solely the responsibility of the authors and does not necessarily represent the official views of PCORI, PCORI's Board of Governors, PCORI's Methodology Committee, NLM, or NHLBI.

\bibliography{references}

%%%%%%%%%%%%%%%%%%%%%%%%%%%%%%%%%%%%%%%%%%%%%%%%%%%%%%%%%%%%
%\input{checklist}

\clearpage
\appendix
\thispagestyle{empty}

\onecolumn
\aistatstitle{Robust Estimation of Heterogeneous Treatment Effects in Randomized Trials Leveraging External Data: Supplementary Materials}

\section{PROOFS AND DERIVATIONS} \label{app:proofs}

\subsection{Proof of Theorem~\ref{thm:robustness_pseudo_outcome}}\label{app:robustness_proof}

\begin{proof}
    We write $\psi_{\text{fixed}}=\psi(O;\eta_{\text{fixed}})$.
    
    First, we need to prove conditional unbiasedness, $\E[\psi_{\text{fixed}} \mid X=x, S=1] = \tau(x)$, as follows:
\begin{align*}
        \E[\psi_{\text{fixed}} \mid X=x] &= \E\Bigg[ \frac{A}{e(X)}(Y-h_1(X)) - \frac{1-A}{1-e(X)}(Y-h_0(X)) + h_1(X) - h_0(X) 
        \mid X=x,S=1\Bigg] \\
        & = \E\Bigg[ \frac{A}{e(X)}(Y^1-h_1(X)) - \frac{1-A}{1-e(X)}(Y^0-h_0(X)) 
        \mid X=x, S=1\Bigg] + h_1(x) - h_0(x)
\end{align*}
where the second equality follows from consistency in Condition~\ref{asmp:consistency}.
Next, we inspect the first term inside the above expectation, which can be rewritten as follows
\begin{align*}
    \E\Bigg[ \frac{A}{e(X)}(Y^1-h_1(X))\mid X=x, S=1\Bigg] &= \E\Bigg[ \frac{A}{e(X)} \mid X=x, S=1\Bigg] \Bigg(\E\Big[Y^1\mid X=x, S=1\Big] - h_1(x)\Bigg) \\ 
    & = \frac{e(x)}{e(x)}\Bigg(\E\Big[Y^1\mid X=x, S=1\Big] - h_1(x)\Bigg) \\
    & = \E\Big[Y^1\mid X=x, S=1\Big] - h_1(x)
\end{align*}
where the first equality follows from conditional exchangeability in the trial population, $Y^a\indep A \mid X, S=1$, in Condition~\ref{asmp:strong_ignorability_trial} and the second equality follows from that $\E[A\mid X=x, S=1]=e(x)$. Similarly, we can show that 
\begin{equation*}
     \E\Bigg[ \frac{1-A}{1-e(X)}(Y^0-h_0(X))\mid X=x, S=1\Bigg] = \E\Big[Y^0\mid X=x, S=1\Big] - h_0(x)~.
\end{equation*}
Putting all of the above together, we see that 
\begin{align*}
    \E[\psi_{\text{fixed}}\mid X=x] & = \E\Big[Y^1 - Y^0\mid X=x, S=1\Big] = \tau(x)
\end{align*}

Next, we show that $\tau = \argmin_{\tilde\tau \in \mathcal{F}} R(\tilde\tau;\eta_{\text{fixed}})$ when $\tau\in\mathcal{F}$. By adding and subtracting $\tau(X)$ inside $R(\tilde\tau;\eta_{\text{fixed}})$, we can decompose it as
\begin{align*}
    \underbrace{\E\left[(\tau(X) - \tilde\tau(X))^2 \mid S=1\right]}_{(a)} -  \underbrace{\E\left[\left( \tau(X) - \tilde\tau\left(X\right) \right) \left( \psi_{\text{fixed}} - \tau(X)\right) \mid S=1\right]}_{(b)} + \underbrace{\E\left[ \left( \psi_{\text{fixed}}  - \tau(X) \right)^2 \mid S=1\right]}_{(c)}
\end{align*}
First, we see that $(a) = R^*(\tilde\tau)$. Next, we have that $(b)=0$ because
\begin{align*}
    \E&\left[\left(\tau(X) - \tilde\tau\left(X\right) \right) \left( \psi_{\text{fixed}}  - \tau(X)\right) \mid S=1\right] = \\
    & =  \E\left[\E\left[\left(\tau(X) - \tilde\tau\left(X\right) \right) \left( \psi_{\text{fixed}}  - \tau(X)\right) \mid X, S=1 \right]\mid S=1\right] \\
    &=  \E\left[\left(\tau(X) - \tilde\tau\left(X\right) \right) \E\left[ \left( \psi_{\text{fixed}}  - \tau(X)\right) \mid X, S=1 \right]\mid S=1\right] \\
    &= 0
\end{align*}
where the last equality follows from conditional unbiasedness such that $\E\left[ \left( \psi_{\text{fixed}}  - \tau(X)\right) \mid X, S=1 \right]=0$.
Finally, $(c)=C$ is a real-valued constant $C\geq 0$ independent of $\tilde\tau$. Thus, we can write that
\begin{equation*}
    R(\tilde\tau;\eta_{\text{fixed}}) = R^*(\tilde\tau) + C
\end{equation*}
which implies that
\begin{equation*}
\argmin_{\tilde\tau\in\mathcal{F}}R(\tilde\tau;\eta_{\text{fixed}}) = \argmin_{\tilde\tau\in\mathcal{F}} \{R^*(\tilde\tau) + C\} = \argmin_{\tilde\tau\in\mathcal{F}} R^*(\tilde\tau) = \tau~.
\end{equation*}
\end{proof}

\subsection{Decomposition of the Sample Analog Pseudo-risk} \label{app:decomposition}
We have that
\begin{align*}
     \widehat{R}(\tilde\tau;\hat\eta) &= \frac{1}{n_1} \sum_{i:S_i=1} \left(\widehat{\psi}_i - \tilde\tau(X_i)\right)^2 \\
     &= \frac{1}{n_1} \sum_{i:S_i=1} \left(\widehat{\psi}_i - \tau(X_i) + \tau(X_i) - \tilde\tau(X_i)\right)^2 \\
     &= \frac{1}{n_1}\sum_{i: S_i=1} \Bigg\{ \left( \tau(X_i) - \tilde\tau\left(X_i\right) \right)^2- 2\left( \tau(X_i) - \tilde\tau\left(X_i\right) \right) \left( \widehat{\psi}_i - \tau(X_i)\right)+ \left( \widehat{\psi}_i - \tau(X_i) \right)^2\Bigg\}~.
\end{align*}

\subsection{Proof of Lemma~\ref{lem:variance_bound}}\label{app:variance_bound}

\begin{proof}
    To make it more explicit that the estimated nuisance models $\hat\eta$ are obtained independently of the observations used to compute the pseudo-outcomes, we denote it as $\eta_{\text{fixed}}=\{h_1,h_0\}$. Moreover, we write $\psi_{\text{fixed}}=\psi(O;\eta_{\text{fixed}})$. 

    Defining $\epsilon:=\left( \tau(X) - \tilde\tau\left(X\right) \right) \left( \psi_{\text{fixed}} - \tau(X)\right)$, we then have that $\E[\epsilon\mid S=1] = 0$ which we showed in the proof of Theorem~\ref{thm:robustness_pseudo_outcome}. Next, we note that 
    \begin{align*}
        \V(\epsilon \mid S=1) &= \E\left[ \epsilon^2 \mid S=1 \right] \\ 
        & = \E\left[\left( \tau(X) - \tilde\tau\left(X\right) \right)^2 \left( \psi_{\text{fixed}} - \tau(X)\right)^2 \mid S=1 \right]  \\
        &= \E\left[ \E\left[\left( \tau(X) - \tilde\tau\left(X\right) \right)^2 \left( \psi_{\text{fixed}} - \tau(X)\right)^2 \mid X, S=1 \right] \mid S=1 \right] \\
        &= \E\left[ \left( \tau(X) - \tilde\tau\left(X\right) \right)^2  \E\left[\left( \psi_{\text{fixed}} - \tau(X)\right)^2 \mid X, S=1 \right] \mid S=1 \right] \\
        & = \E\left[ \left( \tau(X) - \tilde\tau\left(X\right) \right)^2  \V\left( \psi_{\text{fixed}} \mid X, S=1 \right)\mid S=1 \right]
    \end{align*}
    where the first equality follows from that $\E[\epsilon\mid S=1]=0$ and the last from the conditional unbiasedness of the pseudo-outcome, $\E[\psi_{\text{fixed}} \mid X=x, S=1] = \tau(x)$, which we derived in the proof of Theorem~\ref{thm:robustness_pseudo_outcome}.

    Next, we show how to upper-bound $\V(\epsilon\mid S=1)$ as follows:
    \begin{align*}
        \V(\epsilon\mid S=1) & = \E\left[ \left( \tau(X) -\tilde\tau\left(X\right) \right)^2  \V\left( \widehat{\psi} \mid X, S=1 \right)\mid S=1 \right] \\
        & \leq \tilde{C} \cdot \E\left[\V\left(\psi_{\text{fixed}} \mid X, S=1\right) \mid S=1\right]
    \end{align*}
    where the inequality holds if define the constant $\tilde{C}=\max_{x\in\mathcal{X}}(\tau(x) - \tilde\tau(x))^2 \geq 0$ . Next, we have from the law of total variance that 
    \begin{align*}
    \E\left[\V\left(\psi_{\text{fixed}} \mid X, S=1\right) \mid S=1\right] &= \V(\psi_{\text{fixed}} \mid S=1) -  \V\left(\E\left[ \psi_{\text{fixed}} \mid X, S=1 \right]\mid S=1\right) \\
    & = \V(\psi_{\text{fixed}} \mid S=1) -  \V\left(\tau\left(X\right) \mid S=1\right)~.
    \end{align*}
    where the second equality follows from the conditional unbiasedness of the pseudo-outcomes. Thus, so far, we have $\V(\epsilon\mid S=1) \leq \tilde{C} \left[ \V(\psi_{\text{fixed}} \mid S=1) -  \V\left(\tau\left(X\right) \mid S=1\right)\right]$.
    
    Next, we inspect the variance $\V(\psi_{\text{fixed}} \mid S=1)$, which we rewrite using the law of total variance:
    \begin{align*}
       \V(\psi_{\text{fixed}} \mid S=1) &= \E\left[ \V\left( \psi_{\text{fixed}} \mid Y^1, Y^0, X, S=1 \right)\mid S=1 \right] + \V\left(\E\left[\psi_{\text{fixed}} \mid Y^1, Y^0, X, S=1 \right]\mid S=1 \right) \\
        & = \E\left[ \V\left( \psi_{\text{fixed}} \mid Y^1, Y^0, X, S=1 \right)\mid S=1 \right] + \V\left(Y^1-Y^0\mid S=1 \right)
    \end{align*}
    where the second inequality follows from that 
    \begin{equation*}
    \E\left[\psi_{\text{fixed}} \mid Y^1, Y^0, X, S=1 \right] = \E\left[Y^1-Y^0\mid Y^1, Y^0, X, S=1 \right] = Y^1-Y^0
    \end{equation*} where the first equality stems from the conditional unbiasedness of the pseudo-outcomes.

    We next write $\psi_{\text{fixed}}=\psi_{1,\text{fixed}}-\psi_{0,\text{fixed}}$ where   $\psi_{a,\text{fixed}}=\frac{\mathbf{1}(A=a)}{\mathbf{1}(A=1)e(X) + \mathbf{1}(A=0)(1-e(X))}(Y-h_a(X)) + h_a(X)$. This will help us simplify the expression for the above inner conditional variance as follows,
    \begin{align*}
        \V&\left(\psi_{\text{fixed}} \mid Y^1, Y^0, X, S=1 \right) = \V\left( \psi_{1,\text{fixed}}-\psi_{0,\text{fixed}} \mid Y^1, Y^0, X, S=1 \right) \\
        & = \E\left[ \left\{\psi_{1,\text{fixed}}-\psi_{0,\text{fixed}} - \underbrace{\E\left[\psi_{1,\text{fixed}}-\psi_{0,\text{fixed}} \mid Y^1, Y^0, X, S=1\right]}_{=Y^1-Y^0}\right\}^2\mid Y^1, Y^0, X, S=1 \right]  \\
        & = \E\left[\left\{\left(\psi_{1,\text{fixed}} - Y^1\right) -\left(\psi_{0,\text{fixed}}  - Y^0\right) \right\}^2\mid Y^1, Y^0, X, S=1 \right] \\
        & \leq 2 \E\left[\left(\psi_{1,\text{fixed}}-Y^1\right)^2 +\left(\psi_{0,\text{fixed}} - Y^0\right)^2 \mid Y^1, Y^0, X, S=1 \right]
    \end{align*}
    where the third inequality follows again from the conditional unbiasedness of the pseudo-outcomes and the final inequality from that $(a-b)^2\leq 2(a^2+b^2)$ for any real numbers $a$ and $b$.
    At last, we note that 
    \begin{align*}
        \E&\left[\left(\psi_{1,\text{fixed}}-Y^1\right)^2 \mid Y^1, Y^0, X, S=1 \right] = \\
        &=\E\left[\left(\frac{A}{e(X)}(Y-h_1(X)) + h_1(X)-Y^1\right)^2 \mid Y^1, Y^0, X, S=1 \right] \\
        &=\E\left[\left(\frac{A}{e(X)}(Y^1-h_1(X)) + h_1(X)-Y^1\right)^2 \mid Y^1, Y^0, X, S=1 \right] \\
        & = \E\left[\left(\frac{A}{e(X)} - 1 \right)^2\mid X, S=1 \right] (Y^1-h_1(X))^2  \\
        & = \frac{1-e(X)}{e(X)}(Y^1-h_1(X))^2
    \end{align*}
    where the second equality follows from consistency (Condition~\ref{asmp:consistency}) and the third equality from conditional exchangeability in the trial population (Condition~\ref{asmp:strong_ignorability_trial}). Similarly, we have that 
    \begin{align*}
         \E&\left[\left(\psi_{0,\text{fixed}}-Y^0\right)^2 \mid Y^1, Y^0, X, S=1 \right] = \frac{e(X)}{1-e(X)}(Y^0-h_0(X))^2~.
    \end{align*}
    At last, plugging the above expressions back into our original expression for $\V(\psi_{\text{fixed}} \mid S=1)$, we obtain the inequality
    \begin{align*}
         \V(\epsilon \mid S=1)  &\leq \tilde{C} \Bigg\{2\E\left[\frac{1-e(X)}{e(X)}(Y^1-h_1(X))^2 \mid S=1\right] + 2\E\left[\frac{e(X)}{1-e(X)}(Y^0-h_0(X))^2\mid S=1\right] \\
        &\quad + \V\left( Y^1-Y^0\mid X, S=1 \right) - \V\left(\tau\left(X\right)\mid S=1\right) \Bigg\}  \\
        &= \tilde{C} \Bigg\{ 2\E\left[\frac{1-e(X)}{e(X)}(Y-h_1(X))^2\mid X, A=1, S=1\right] + 2\E\left[ \frac{e(X)}{1-e(X)}(Y-h_0(X))^2\mid  X, A=0, S=1 \right] \\
        &\quad + \V\left( Y^1-Y^0\mid S=1 \right)  - \V\left(\tau\left(X\right)\mid S=1\right) \Bigg\}
    \end{align*}
    where the equality follows from consistency and conditional exchangeability in the trial population again (Condition~\ref{asmp:consistency} and~\ref{asmp:strong_ignorability_trial}).
\end{proof}

\subsection{Proof of Theorem~\ref{thm:variance_rewritten}} \label{app:variance_rewritten}
\begin{proof}
    We begin by writing
    \begin{align*}
        L_a(h) & = \E\left[\left\{\frac{1-e(X)}{e(X)}\right\}^{2a-1}\left(Y-h\left(X\right)\right)^2 \mid A=a, S=1\right] \\
        &= \E\left[ \frac{S}{\Pr(S=1\mid A=a)}\left\{\frac{1-e(X)}{e(X)}\right\}^{2a-1}\left(Y-h\left(X\right)\right)^2 \mid A=a\right]\\
        &= \E\left[ \E\left[ \frac{S}{\Pr(S=1\mid A=a)}\left\{\frac{1-e(X)}{e(X)}\right\}^{2a-1}\left(Y-h\left(X\right)\right)^2 \mid X, A=a \right] \mid A=a\right]\\
        &= \E\left[\frac{1}{\Pr(S=1\mid A=a)}\left\{\frac{1-e(X)}{e(X)}\right\}^{2a-1} \E\left[S \left(Y-h\left(X\right)\right)^2 \mid X, A=a \right] \mid A=a\right]~.
    \end{align*}
    We inspect the inner conditional expectation, $\E\left[S \left(Y-h\left(X\right)\right)^2 \mid X, A=a \right]$, and note that 
    \begin{align*}
        \E\left[S \left(Y-h\left(X\right)\right)^2 \mid X, A=a \right] &= \E[S\mid X, A=a] \E[\left(Y-h\left(X\right)\right)^2 \mid X, A=a] \\
        &=\Pr(S=1\mid X, A=a) \E[\left(Y-h\left(X\right)\right)^2 \mid X, A=a] 
    \end{align*}
    where the first equality follows from that $Y\indep S\mid (X, A)$ holds under Conditions~\ref{asmp:consistency}-\ref{asmp:transportability}. To show this, we note that the conditional independencies $Y^a\indep A\mid (X, S=1)$ (Condition~\ref{asmp:strong_ignorability_trial}) and $Y^a\indep A\mid (X, S=0)$ (Condition~\ref{asmp:strong_ignorability_external}) jointly imply that $Y^a\indep A \mid (X,S)$. Combining this conditional independence statement with $Y^a\indep S\mid X$ from Condition~\ref{asmp:transportability}, they together imply $Y^a \indep (A,S)\mid X$. Thus, from the weak union of conditional independence, we have that $Y^a\indep S\mid (X,A)\Rightarrow Y\indep S\mid (X,A)$ where the final implication follows from consistency (Condition~\ref{asmp:consistency}).

    Combining all of the above, we finally obtain the following expression,
    \begin{equation*}
        L_a(h) = \E\left[\frac{\Pr(S=1\mid X, A=a)}{\Pr(S=1\mid A=a)}\left\{\frac{1-e(X)}{e(X)}\right\}^{2a-1} \left(Y-h\left(X\right)\right)^2\mid A=a\right]
    \end{equation*}
\end{proof}

\subsection{Proof of Lemma~\ref{lem:oracle_weight_combined}}\label{app:oracle_weight_combined}

\begin{proof}
The inequality $R^*(\lambda^{\star}) \le \min\{R^*_{QR}, R^*_{DR}\}$ follows directly from the definition of $\lambda^{\star}$ as the minimizer of $R^*(\lambda)$ over the set $\Lambda$. Since both $\lambda=0$ and $\lambda=1$ are in this set, the minimum value $R^*(\lambda^{\star})$ cannot be greater than the value at either endpoint: $R^*(\lambda^{\star}) \le R^*(0) = R^*_{DR}$ and $R^*(\lambda^{\star}) \le R^*(1) = R^*_{QR}$. Therefore, $R^*(\lambda^{\star}) \le \min\{R^*_{DR}, R^*_{QR}\}$.

For uniqueness, we note that the risk $R^*(\lambda)$ can be expanded as
\begin{align*}
  R^*(\lambda) &= \E\bigl[ (V+\lambda(U-V))^{2} \mid S=1 \bigr] \\
             &= \E[(U-V)^2] \lambda^2 + 2\E[V(U-V)] \lambda + \E[V^2] \\
             &= A\lambda^2 + B\lambda + C,
\end{align*}
where $U=\hattauQR(X)-\cate(X)$, $V=\hattauDR(X)-\cate(X)$, $A=\E[(U-V)^2]$, $B=2\E[V(U-V)]$, and $C=\E[V^2]$. If $A > 0$, the risk function $R^*(\lambda)$ is a strictly convex quadratic function. A strictly convex function has a unique minimum over any convex set, including the set $\Lambda$. Therefore, if $A>0$, the minimizing weight $\lambda^{\star}$ is unique.
\end{proof}

\subsection{Proof of Theorem~\ref{thm:cv_risk_combined}}

\begin{proof}
   % TODO: Define the risk \widetilde R(\lambda) and justify its relation to R^*(\lambda), possibly via a Proposition like the original proof mentioned.
   Let $\widetilde R(\lambda)$ be the expectation of the cross-validated loss term.
Since the squared loss is bounded in $[0,B]$,  $\Lambda:=[0,1]$ is a compact subset of $\mathbb R$, the uniform law of large numbers gives
   \[
\sup_{\lambda\in \Lambda}\bigl|\widehat R_{CV}(\lambda)-\widetilde R(\lambda)\bigr|
     \;=\;o_p(1).
   \]
Further, by definition $\widehat R_{CV}(\hat\lambda^\star)\le\widehat R_{CV}(\lambda^\star)$, hence
   \[
     \widetilde R(\hat\lambda^\star)
     =\widehat R_{CV}(\hat\lambda^\star)+o_p(1)
     \;\le\;
     \widehat R_{CV}(\lambda^\star)+o_p(1)
     =\widetilde R(\lambda^\star)+o_p(1).
   \]
Finally, it follows from the proof in Lemma~\ref{lem:variance_bound} that $\widetilde R(\lambda)=R^*(\lambda)+C$ for some constant $C$ 
   \[
     R^*(\hat\lambda^\star)+C
     \;\le\;
     R^*(\lambda^\star)+C
     +o_p(1)
     \quad\Longrightarrow\quad
     R^*(\hat\lambda^\star)\;\le\;R^*(\lambda^\star)+o_p(1).
   \]
Finally, Lemma~\ref{lem:oracle_weight_combined} yields
   \[
     R^*(\lambda^\star) 
     \;\le\;\min\{R^*_{DR}, R^*_{QR}\},
   \]
   completing the chain:
   \[
     R^*(\hat\lambda^\star)
     \;\le\;
     R^*(\lambda^\star)
     \;+\;o_p(1)
     \;\le\;
     \min\{R^*_{DR}, R^*_{QR}\}
     \;+\;o_p(1).
   \]
\end{proof}

\section{CONNECTIONS TO R-LEARNER}
\label{app:R-learner}
It can be shown that the DR-learner~\citep{kennedy2023towards} and the R-learner~\citep{nie2021quasi} both minimize two closely related loss functions when estimating the CATE. Below, we outline this connection and discuss its implications for connections between our proposed QR-learner and the R-learner.

\citet{morzywolek2023weighted} introduced a general formulation of a loss function,
\begin{equation}
    R_o(\tilde\tau;\eta, \lambda) = \frac{1}{\E[\lambda\{e(X)\}]}\E\left[\rho\{A,e(X);\lambda\} \{ \phi(O;\eta, \lambda) - \tilde\tau\}^2 \mid S=1 \right]
\end{equation}
which, when minimized, yield a broad class of CATE learners. Here, the function $\rho\{A,e(X);\lambda\}$ and pseudo-outcome $\phi(O;\eta, \lambda)$ are defined as
\begin{align*}
    \rho\{A,e(X);\lambda\} & = \{A-e(X)\}\lambda'\{e(X)\}+\lambda\{e(X)\}\\
    \phi(O;\eta, \lambda) & = \frac{\lambda\{e(X)\}}{\rho\{A,e(X)\}}\frac{A-e(X)}{e(X)(1-e(X))}(Y-g_A(X)) + g_1(X)-g_0(X)
\end{align*}
where the nuisance models are $\eta=\{e(X), g_1, g_0\}$ and $g_a=\E[Y\mid X, A=a, S=1]$ for $a\in{0,1}$.

The loss $R_o(\tilde\tau;\eta, \lambda)$ is indexed by a weighting function $\lambda : [0,1]\rightarrow \mathbb{R}$ and different CATE learners are obtained depending on this weighting function. For example,  the DR-learner and R-learner can both be recovered as special cases with different functions $\lambda$.

If we let $\lambda\{e(X)\}\equiv c$ for some constant $c\in\mathbb{R}$, we obtain $\rho\{A,e(X);\lambda\}=c$ and $\phi(O;\eta,\lambda)=\frac{A-e(X)}{e(X)(1-e(X))}(Y-g_A) + g_1(X)-g_0(X)$ which is equivalent to the pseudo-outcome $\psi(O;\eta=\{g_1,g_0\})$ from~\eqref{eq:pseudo_outcome}. Consequently, the loss function $R_0(\tilde\tau;\eta,\lambda)$ then equals the pseudo-risk $R(\tilde\tau;\eta=\{g_1,g_0\})$ of the DR-learner.

Meanwhile, when letting $\lambda\{e(X)\}=e(X)(1-e(X))$ and following some algebraic steps, we arrive at another expression for the loss function $R_0(\tilde\tau;\eta,\lambda)$, given by
\begin{equation*}
     \frac{1}{\E[e(X)(1-e(X))\mid S=1]}\E\left[(\left\{Y-q(X)\right\} - \left\{A-e(X)\right\} \tilde\tau(X))^2\mid S=1\right]~,
\end{equation*}
where $q(X)=\E[Y\mid X, S=1]=e(X)\cdot g_1(X) + (1-e(X))\cdot g_0(X)$. This expression can be recognized as the loss function of the R-learner~\citep{nie2021quasi}.

The above observations highlight an interesting point: the difference between the DR-learner and R-learner depends on the propensity score $e(X)$. If $e(X)$ is constant, then $\lambda\{e(X)\}=e(X)(1-e(X))$ is also constant, and the two loss functions coincide. Meanwhile, if $e(X)$ is non-constant, the DR-learner and R-learner lead to different CATE learners.

This connection has direct implications for our proposed QR-learner. As discussed in the main paper, the QR-learner has natural links to the DR-learner, since it uses the same loss function but fits the nuisance models differently with the help of external data. When the propensity score $e(X)$ is constant, a scenario realistic in randomized trials which often have fixed treatment probabilities, the R-learner and DR-learner coincide. Hence, in these cases, we argue that the QR-learner can be viewed as minimizing an analogous loss function to either the DR-learner or the R-learner. On the other hand, if the treatment probabilities are covariate-dependent and not fixed, the QR-learner does not minimize the same loss as the R-learner, but rather that of the DR-learner.

\section{PSEUDO-CODE FOR QR-LEARNER}\label{app:pseudo_code}

\begin{algorithm}[h!]
\caption{QR-learner: Quasi-optimized Randomization-aware Learner}
\label{alg:qr_learner}
\begin{algorithmic}[1]
\REQUIRE Data $\mathcal{D}=\{(X_i,S_i, A_i, Y_i)\}_{i=1}^n$, treatment propensity score $e(X)$, model classes $\mathcal{H}$ (for outcome models), $\mathcal{G}$ (for treatment participation probability)  and $\mathcal{F}$ (for CATE), number of folds $K$ 

\STATE Partition $\mathcal{D}$ into $K$ folds $\{\mathcal{D}^k\}_{k=1}^K$, stratified by treatment $A$ and study indicator $S$.

\FOR{$k = 1$ to $K$}
    \STATE \textbf{Stage 1: Estimate nuisance models}
    
    \FOR{$a \in \{0, 1\}$}
        \STATE Define $\mathcal{D}_a^k = \{i \in \mathcal{D}^k : A_i = a\}$
        \STATE Estimate $\Pr(S=1 \mid X, A=a)$ using $\mathcal{D}_a^k$ using an estimator $\widehat{\pi}_a(X)$ from model class $\mathcal{G}$
        \STATE Solve optimization problem in~\eqref{eq:optimization_h}
    \ENDFOR
    
    \STATE Set $\hat{\eta}^{(k)} = \{\hat{h}_1^{(k)}, \hat{h}_0^{(k)}\}$
    
    \STATE \textbf{Stage 2: Estimate CATE model}
    
    \STATE Let $\mathcal{D}^{-k} = \mathcal{D} \setminus \mathcal{D}^k$
    
    \STATE Compute pseudo-outcome $\psi(O_i; \hat{\eta}^{(k)})$ for each $i \in \mathcal{D}^{-k}$ according to~\eqref{eq:pseudo_outcome}
    
    \STATE Estimate CATE on trial data by solving $\hat{\tau}^{(k)} = \argmin_{\tilde{\tau} \in \mathcal{F}} \sum_{i \in \mathcal{D}^{-k} : S_i = 1} \bigl(\psi(O_i; \hat{\eta}^{(k)}) - \tilde{\tau}(X_i)\bigr)^2$
    
\ENDFOR

\RETURN $\hat{\tau}(X) = \frac{1}{K} \sum_{k=1}^{K} \hat{\tau}^{(k)}(X)$
\end{algorithmic}
\end{algorithm}

\section{STATISTICAL INFERENCE WITH QR-LEARNER}
\label{app:QR_learner_guarantees}

In this section, we discuss when statistical inference is feasible for the estimated CATE function $\hat\tau(x)$ produced by the QR-learner. Although a full theoretical analysis is beyond the scope of this work, we outline key conditions under which inference is expected to be valid.

Under standard regularity conditions and with the use of cross-fitting, inference on $\hat\tau(x)$ becomes possible when the regression of pseudo-outcomes $\widehat{\psi}$ on covariates $X$ is performed using a low-dimensional model~\citep[Chapter 14]{chernozhukov2024}. This is enabled by the Neyman orthogonality of the pseudo-outcomes in~\eqref{eq:pseudo_outcome}, which ensures that the impact of nuisance estimation errors on $\hat\tau(x)$ is second-order~\citep{foster2023orthogonal}.

Moreover, \citet{kennedy2023towards} show that if a ``stable regressor'' is used to estimate $\tau(x)$ in the DR-learner (we refer to their paper for the formal definition), statistical inference may also be guaranteed. Examples of such regressors include linear regression, smoothing splines, and kernel ridge regression. Importantly, this remains feasible even if the nuisance components $\hat\eta$ converge at slower rates than $\hat\tau(x)$, since their influence appears as a product of estimation errors -- specifically, those from $\hat\eta$ and the estimated propensity score (if one were to estimate it). In randomized trials, where the true propensity score is known, we could expect this product to vanish rapidly, making the impact of nuisance estimation asymptotically negligible on the final CATE estimate $\hat\tau(x)$. In this case, asymptotically valid inference for $\hat\tau(x)$ may be warranted as long as the regressor used for the final stage regressor has inferential guarantees~\citep{kennedy2023towards}.

\section{DISCUSSION ON METHODS FROM ASIAEE ET AL.}\label{app:asiaee}

In this section, we explain the method R-OSCAR (Robust Observational Studies for CMO-Augmented RCT) from~\citet{asiaee2023improving} as well as another variant proposed in an earlier version of their manuscript (see~\citet{asiaee2023leveraging}), refered to as CFACE (CounterFactual Average Covariate Effect). We start with explaining CFACE since it came out first.

\paragraph{CFACE}
The central idea of CFACE is to estimate a trial-specific CATE learner leveraging pseudo-outcomes which fits into our randomization-aware framework, but using a different procedure for estimating the outcome nuisance models. Specifically, we instead consider randomization-aware pseudo-outcomes on the restricted form:
\begin{equation}
\psi(O; \eta=\{m,m\}) = \frac{A-e(X)}{e(X)(1-e(X))}(Y-m(X)),
\end{equation}
where the propensity score $e(X)$ is known and the nuisance models are constrained to be identical, $h_1=h_0=m$. Then,~\citet{asiaee2023leveraging} show that $m$ can be chosen to minimize the conditional variance,
\begin{align*}
m^*(x) &= \argmin_m \V\left( \psi(O; \eta=\{m,m\}) \mid X=x, S=1\right) \\
&= e(x)\cdot \mu^{S=1}_0(x) + (1-e(x))\cdot \mu^{S=1}_1(x),
\end{align*}
with $\mu^{S=1}_a(x) = \mathbb{E}[Y \mid X = x, A = a, S = 1]$. Under transportability (Condition 4), $\mu^{S=1}_a(x)$ also equals $\mu^{S=0}_a(x)=\mathbb{E}[Y \mid X = x, A = a, S = 0]$ and can thus be estimated from external data. Thus, we can fit $\mu^{S=0}_a(x)$ on only the external data for $a \in \{0,1\}$ and then compute pseudo-outcomes using $m^*$ following a two-stage procedure similar to the DR- and QR-learners. However, in their updated manuscript, the authors acknowledge that this procedure is possible but caution that it may fail when trial and external populations are misaligned (i.e., when Condition~4 does not hold). For this reason, they now recommend against this and instead focus on a new CATE learner, called R-OSCAR.

\paragraph{R-OSCAR}
R-OSCAR departs from CFACE by proposing to fit $\mu^{S=0}_a(x)$ on external data and correct for possible transportability violations by modeling the difference
\begin{equation}
\delta_a(x) = \mathbb{E}[Y \mid X = x, A = a, S = 1] - \mathbb{E}[Y \mid X = x, A = a, S = 0].
\end{equation}
Since their focus is on the CATE itself, they further propose modeling the direct discrepancy in the CATE between the populations:
\begin{equation}
\delta(x) = \mathbb{E}[Y^1-Y^0 \mid X=x, S=1] - \mathbb{E}[Y^1-Y^0 \mid X=x, S=0].
\end{equation}
Using the optimality of $m^*$, computed using estimates of $\mu^{S=0}_a$ and $\delta_a$, and the unbiasedness of $\psi(O;\eta=\{m^*,m^*\})$ for the true CATE, they derive a procedure to also estimate $\delta(x)$. 
Their final CATE learner, R-OSCAR, is then
\begin{equation}
\hat\tau_{\text{R-OSCAR}}(x) = (\hat\mu^{S=0}_1(x)+\hat\delta_1(x)) - (\hat\mu^{S=0}_0(x)+\hat\delta_0(x)) + \hat\delta(x),
\end{equation}
where $\hat\mu^{S=0}_a$, $\hat\delta_a$, and $\hat\delta$ are estimated using trial and external data. Modeling the differences $\delta_a$ and $\delta$ resembles the additive bias correction of~\citet{kallus2018removing} more than our approach.
In fact, R-OSCAR could be seen as a T-learner with additive bias correction, and diverges substantially from the two-stage procedures of the DR- and QR-learners. By contrast, the implementation of CFACE follows a two-stage design and aligns more closely with our framework. 

\section{EXPERIMENTAL DETAILS} \label{app:experiment}

\subsection{Data-Generating Process in Simulation Studies}

We simulate data as follows: We set $S_i=1$ for $i=1,\dots,n_1$, and for $S_i=0$ for $i=n_1+1,\dots, n_1+n_0$, and sampled a Normal $d$-dimensional covariate according to $X_i\sim N(\mu_{S_i}, \frac{1}{\sqrt{d}}\Sigma)$ with the mean $\mu_1=\mathbf{0}$ or $\mu_0=0.2\cdot \mathbf{1}$ and the covariance matrix $\Sigma$ of shape $d\times d$ had its diagonal elements set to $1$ and its off-diagonal elements set to $0.1$. 
Thereafter, we sampled the treatment $T_i\sim\text{Bern}(e(X_i,S_i))$ according to the Bernoulli probability 
\begin{equation*}
    e(X_i,S_i) = 
    \begin{cases}
        0.5,& \text{ if }S_i=1 \\
        \frac{1}{1+\exp{\{-(\alpha_0+\alpha^\top X_i)}\}}, & \text{otherwise}
    \end{cases}
\end{equation*}
Finally, we computed outcomes $Y_i=b(X_i)+A_i\cdot \tau(X_i)+\varepsilon_i$ where the noise variables were sampled according to $\varepsilon_i\sim N(0,\sigma^2=1/4)$. 

For the experiment in Section~\ref{sec:simulation_rmse}, we modeled a highly non-linear baseline risk together with a linear CATE, which as done by defining:
\begin{align*}
    b(X_i) &= \sum_{j=1}^d \frac{3}{d}\cos \left(\frac{3}{2}X_{ij}\right) + \sum_{j=1}^d \sum_{j'=1}^d \frac{1}{d}X_{ij}X_{ij'}, \\
    \tau(X_i) &= \sum_{j=1}^d  \frac{1}{d}X_{ij}.
\end{align*}
In the scenario where Conditions~\ref{asmp:strong_ignorability_external} and~\ref{asmp:transportability} held, we set the covariate dimension to $d=5$. To simulate violations of these assumptions, we increased the covariate dimension to $d=7$ but masked the last two dimensions, so that only 5 covariates remained observed. 

For the experiment in Section~\ref{sec:simulation_power}, we considered a setting with a linear baseline outcome and a sparse linear CATE that depended only on a single covariate. To violate transportability (Condition~\ref{asmp:transportability}), this time we encoded different functions $\tau(X)$ for the trial and external population. Specifically, we defined:
\begin{align*}
b(X_i) &= \sum_{j=1}^d \frac{1}{d} X_{ij}, \\
\tau(X_i) &= \begin{cases}
    \beta \cdot d \cdot X_{i1}, &\text{ if }S_i=1\\
    \left(\beta + \frac{1}{20}\right) \cdot d \cdot X_{i1}, & \text{ otherwise}
\end{cases}
\end{align*}
where $\beta$ was a tunable parameter that controlled the size of the interaction effect between the treatment $A_i$ and the first covariate $X_{i1}$. As before, we set $d=5$.

\subsection{Implementation Details For CATE Learners} \label{app:implementation_details}
For the experiments in Section~\ref{sec:simulation_rmse} and~\ref{sec:star}, we implemented the CATE learners as follows.

For the estimators used inside the CATE learners we used implementations from the \textit{scikit-learn} Python package~\citep{scikit-learn} . For the DR-learner, T-learner, pooled T-learner, CFACE, and QR-learner, we used histogram-based gradient boosting regression tree using default hyperparameters. As the final CATE regressor in the two-stage CATE learners (all of the above except the T-learner variants), we used a linear regression model. For estimating $\pi_a(X)=\Pr(S=1\mid X,A=a)$, we used a cross-validated logistic regression with ridge penalty. For KSP, we fitted the DR-learner on the external dataset and then used a linear regression to estimate the bias model. For R-OSCAR, we also used histogram-based gradient boosting regression for the outcome nuisance models and a cross-validated linear regression with ridge penalty for estimating both its bias functions. We applied cross-fitting to all two-stage CATE learners using two folds consistently. For cross-fold validation in the combined learner, we used three folds. 

To predict with the average treatment effect (ATE), we used the difference-in-means estimate $$\hat\tau_{DM}=\frac{\sum_{i=1}^{n_1} A_iY_i}{\sum_{i=1}^{n_1} A_i} - \frac{\sum_{i=1}^{n_1} (1-A_i)Y_i}{\sum_{i=1}^{n_1} 1-A_i}$$ as a constant CATE prediction $\hat\tau(x)=\hat\tau_{DM}$ for all $x$.

For the case study Tennessee STAR dataset, we made a few changes to the implementation of all learners. First, due to having a large number of features relative to the sample size, we use ridge-penalized linear regression for the CATE model, and further changed from the default hyperparameters of the histogram-based gradient boosting regression to a max depth of 3 for the decision trees (default is unconstrained depth) and a minimum sample size per leaf of 5 (default is 20). Finally, we used 10 folds for the cross-validation in the combined learner, to prevent the training splits from becoming very small in size.

\subsection{Statistical Tests for Treatment Effect Heterogeneity} \label{app:statistical_tests}

Below we describe our implementations for the statistical tests used to detect treatment effect modification in the experiment in Section~\ref{sec:simulation_power}.

For covariate adjustment, we fit a linear regression model of $Y$ on the covariates $(A, X_1, A \cdot X_1)$ using the trial data. We then estimated 95\% confidence intervals for the coefficient of $A \cdot X_1$ to assess whether it was significantly different from zero. For the pooled covariate adjustment, we followed the same approach but fit the linear regression model on the combined dataset consisting of both the trial data and the external data.

For the DR-learner, QR-learner, and CFACE, we split the data into two folds (stratified by the treatment $A$ and the study population $S$). We used the first fold to estimate the nuisance components in $\eta$ via linear regression using their respective strategies as outlined in the main paper, then computed pseudo-outcomes on the second fold. These pseudo-outcomes were regressed on the feature $X_1$ using another linear regression model, resulting in the first CATE estimate $\hat{\tau}(X_1) = \hat{\alpha}^{(1)} \cdot X_1 + \hat{\beta}^{(1)}$. To utilize the entire dataset efficiently, we repeated the process by swapping the folds, obtaining a second CATE estimate $\hat{\tau}(X_1) = \hat{\alpha}^{(2)} \cdot X_1  + \hat{\beta}^{(2)}$. We then computed 95\% confidence intervals to test the null hypothesis that $\alpha=0$ as follows:
$$\left[\bar{\alpha} - 1.96 \cdot \text{se}(\bar{\alpha}), \; \bar{\alpha} + 1.96 \cdot \text{se}(\bar{\alpha})\right],$$
where $\bar{\alpha} = \frac{1}{2}(\hat{\alpha}^{(1)} + \hat{\alpha}^{(2)})$ and using the individual standard errors $\text{se}(\hat{\alpha}^{(1)})$ and $\text{se}(\hat{\alpha}^{(2)})$ to compute
$$
\text{se}(\bar{\alpha}) = \sqrt{\frac{1}{4} \left( \text{se}(\hat{\alpha}^{(1)})^2 + \text{se}(\hat{\alpha}^{(2)})^2 \right)},
$$
assuming normality and that the covariance between $\hat{\alpha}^{(1)}$ and $\hat{\alpha}^{(2)}$ to be negligible.

\subsection{Tennessee STAR Dataset}\label{app:star}

The data from the STAR study consists of 4218 students: 2811 from rural schools and 1407 from urban schools. The treatment is class size: small ($A = 0$) versus regular ($A = 1$), and the continuous outcome $Y$ is the student's average test score. Our observed covariates include gender, race, birth date, teacher ID, and free lunch eligibility. After applying one-hot encoding to these variables, which are either binary or categorical, we obtain a 310-dimensional feature vector. 

To construct two datasets from different populations, we split the dataset into rural and urban school. One school location is assigned as the target trial population and the other as the external population. In the experiment, we sample without replacement half of the target trial population as a held-out test set, and then sample without replacement $n_1$ and $n_0$ observations from the target trial and external populations, respectively. We kept $n_0=1000$ fixed while varying $n_1$ from 100 to 700 (about half of the number of urban schools).

\begin{figure}[t]
    \centering
    \begin{subfigure}[b]{0.4\linewidth}
        \centering    \includegraphics[width=\linewidth]{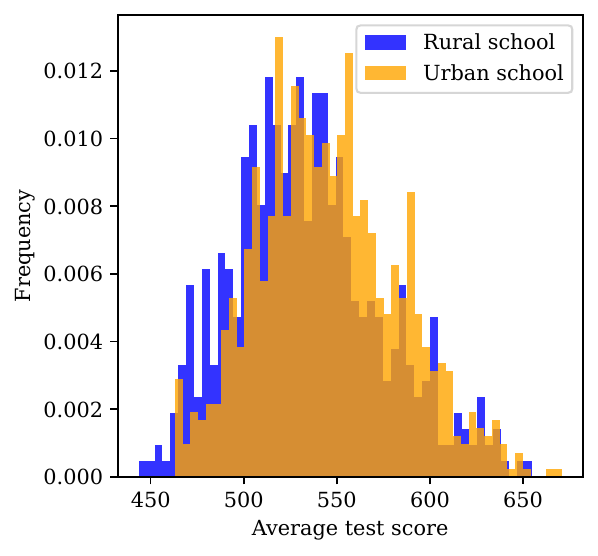}
        \caption{}
        \label{fig:star_outcome_dist}
    \end{subfigure}
    ~
    \begin{subfigure}[b]{0.4\linewidth}
        \centering    \includegraphics[width=\linewidth]{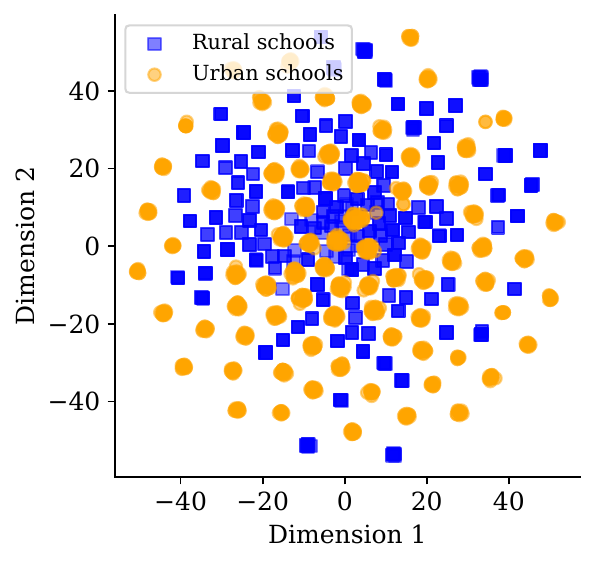}
        \caption{}
        \label{fig:star_tsne}
    \end{subfigure}
    \caption{\textbf{(a)}: Distributions of the outcome (average test scores) for rural and urban schools in the STAR dataset. We observe a slight shift in the mean between the two groups, suggesting potential violations of transportability, as the primary difference between the trial and the external population lies in school location. \textbf{(b)}: t-SNE plot over features colored by study population. We observe some lack of overlap between populations.
    }
    \label{fig:star_visualization}
\end{figure}

We further evaluate whether transportability (Condition~\ref{asmp:transportability}) is violated between the two populations in this dataset. As shown in Figure~\ref{fig:star_outcome_dist}, we can see that the distribution of outcomes (average test scores) differs between rural and urban schools. This may indicate that transportability also is violated if the school location is omitted from the observed covariates $X$. We performed a conditional independence test of $Y\indep S \mid (X,A)$, which is a testable implication for Conditions~\ref{asmp:consistency} to~\ref{asmp:transportability} jointly~\citep{dahabreh2024using}. We applied both a partial linear correlation test and the randomized conditional correlation test (RCoT)~\citep{strobl2019approximate}, implemented in the Python package \textit{pybnesian}~\citep{Atienza2022Pybnesian}. Both tests indicate that the conditional independence is violated at the 5\% significance level, suggesting that one of Conditions~\ref{asmp:consistency} to~\ref{asmp:transportability} may not hold. Since consistency (Condition~\ref{asmp:consistency}) and strong ignorability in both populations (Conditions~\ref{asmp:strong_ignorability_trial} and~\ref{asmp:strong_ignorability_external}) are expected to hold, it is most likely that transportability (Condition~\ref{asmp:transportability}) is violated.

We further used t-SNE for dimensionality reduction to visualize the distribution of covariates across the two populations. These visualizations, shown in Figure~\ref{fig:star_tsne}, reveal limited overlap in certain regions between rural and urban schools.

\subsection{Compute Resources Used for Experiments} \label{app:compute_resources} 

All experiments were run on a CPU machine. The simulations in Sections~\ref{sec:experiment} and~\ref{sec:star} each completed in under 72 hours on a laptop with a 2 GHz Quad-Core Intel Core i5 processor and 16GB of RAM. Including preliminary experiments not shown in the paper, no single run exceeded this runtime.

\section{ADDITIONAL EXPERIMENTAL RESULTS} \label{app:additional_experiments}

\subsection{Analysis of Prediction Bias} 

As a complement to Table~\ref{tab:n1_250_rmse}, which reports the average root mean squared error of each method in our simulation study, Table~\ref{tab:n1_250_bias} presents the corresponding average prediction bias, defined as $\text{bias}(\hat\tau)=\E[\hat\tau(X) - \tau(X) \mid S=1]$. We find that all methods, except the pooled T-learner, were largely unbiased. This aligns with expectations, as we fitted linear models for the CATE which matches with the true underlying linear CATE function. The pooled T-learner, however, exhibited increasing bias when conditions 3 and 4 were violated, which is consistent with its reliance on these conditions for identification of the~CATE.

\begin{table}[t]
    \centering 
    \caption{\small Average prediction bias reported over 500 repeated runs from the simulation study with a trial sample size $n_1=250$ under different scenarios.}
    \label{tab:n1_250_bias}
    \resizebox{0.98\linewidth}{!}{%
    \input{tables/n1_250_bias}
    }
\end{table}

\subsection{Tennessee STAR Dataset} 

\begin{figure*}[h]
    \centering
    \begin{subfigure}[b]{0.45\linewidth}
        \centering    \includegraphics[width=\linewidth]{camera_ready_figures/star_target_urban.pdf}
        \caption{Urban school as target population}
        \label{fig:app_star_urban_target}
    \end{subfigure}
    ~
    \begin{subfigure}[b]{0.45\linewidth}
        \centering
        \includegraphics[width=\linewidth]{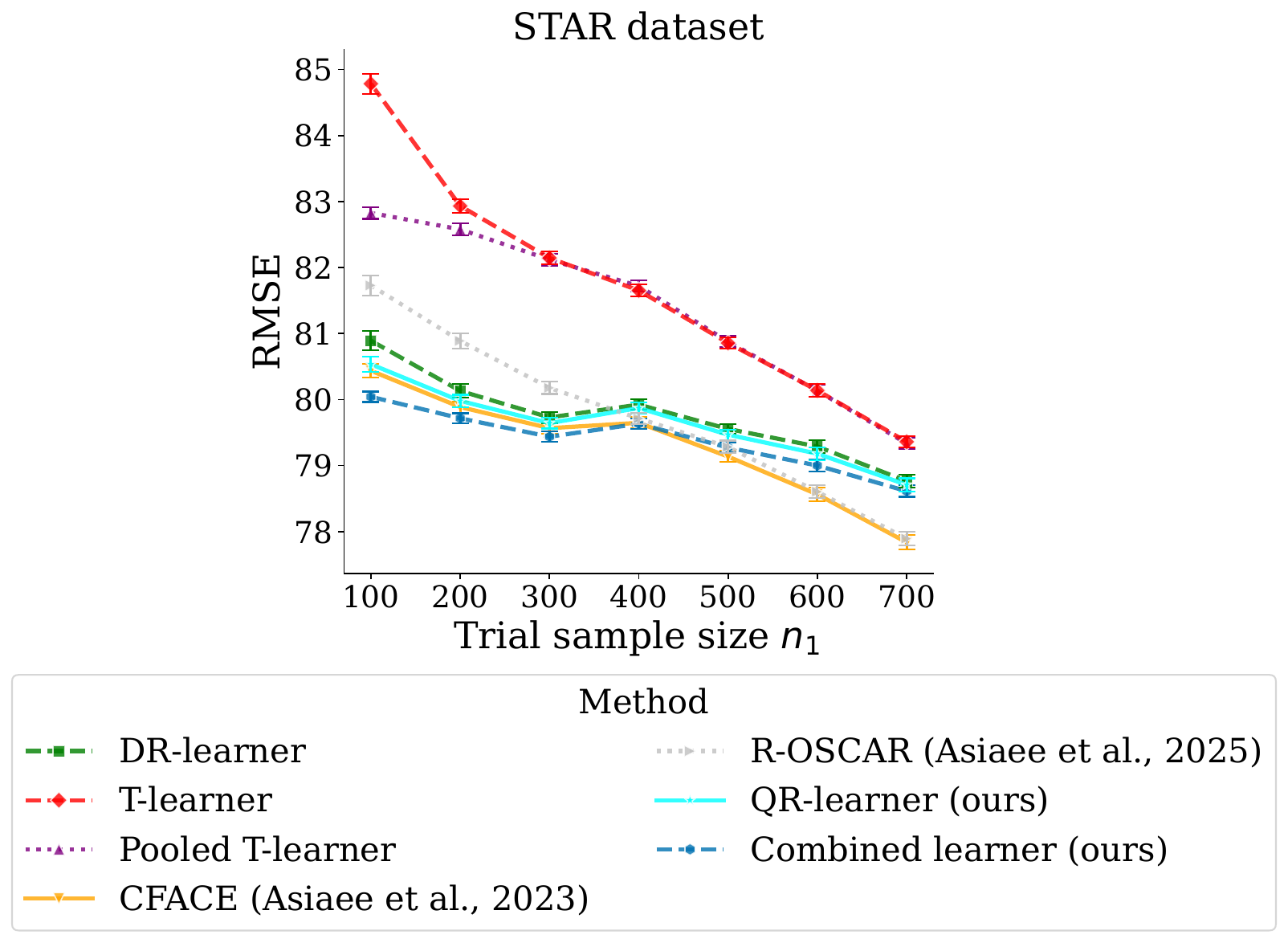}
        \caption{Rural schools as target population}
        \label{fig:app_star_rural_target}
    \end{subfigure}

    \caption{\small We evaluate the RMSE on the STAR dataset when increasing the trial sample size with a fixed external sample size of $n_0=1000$. We report the average RMSE and standard error over 200 repeated runs.}
    \label{fig:app_star}
\end{figure*}

We include both settings from the Tennessee STAR dataset case study. In the first setting, presented in the main paper, we treat the urban schools as the target trial population and the rural schools as the external population. We then flipped the populations, treating the rural schools as the target trial population. Overall, we observe similar trends in both cases, but we highlight below the differences that arise when flipping the two populations. The results from both settings are shown in Figure~\ref{fig:app_star}.

First, we observe that the trial-only and pooled T-learner improve faster when the urban schools are the target population compared to when the rural schools are the target. Second, CFACE and R-OSCAR perform better when the rural schools are the target population. Finally, the QR-learner shows little improvement over the DR-learner when the urban schools are the target, but it exhibits noticeable improvement when the rural schools are the target.

\begin{figure*}[h]
    \centering
        \begin{subfigure}[b]{0.3\linewidth}
        \centering    \includegraphics[width=\linewidth]{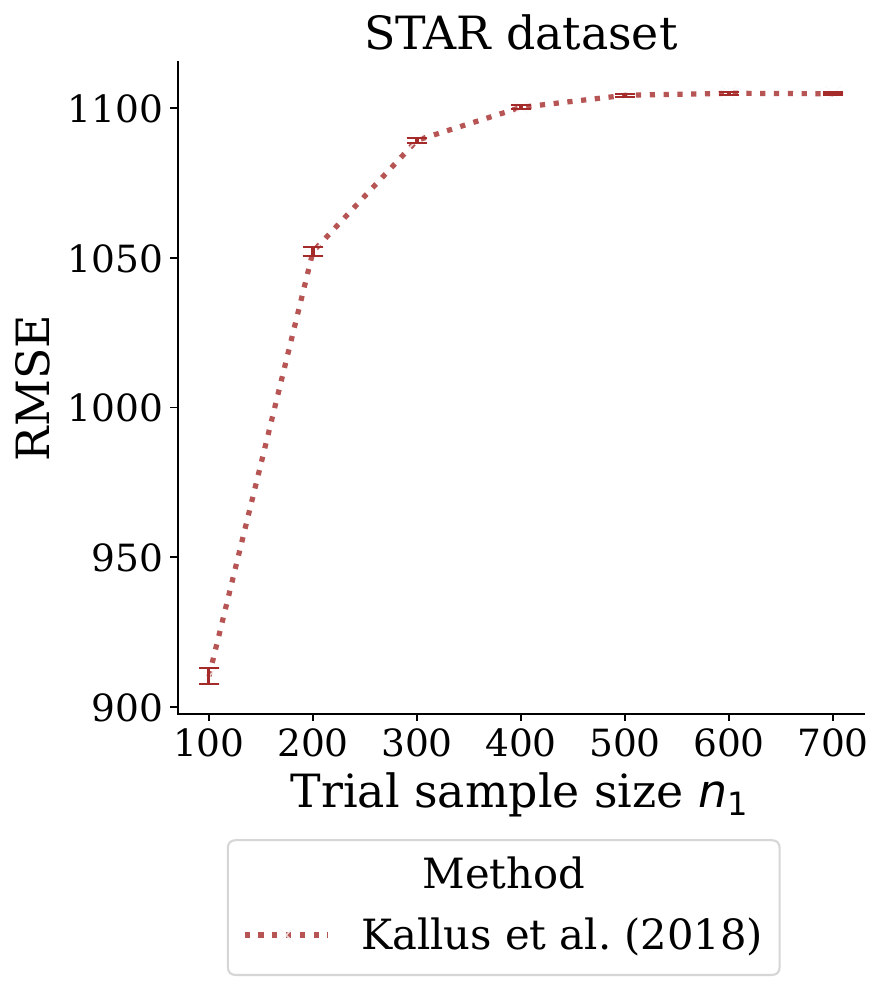}
        \caption{Urban school as target population}
        \label{fig:app_star_urban_target_kallus}
    \end{subfigure}
    ~
    \begin{subfigure}[b]{0.3\linewidth}
        \centering
        \includegraphics[width=\linewidth]{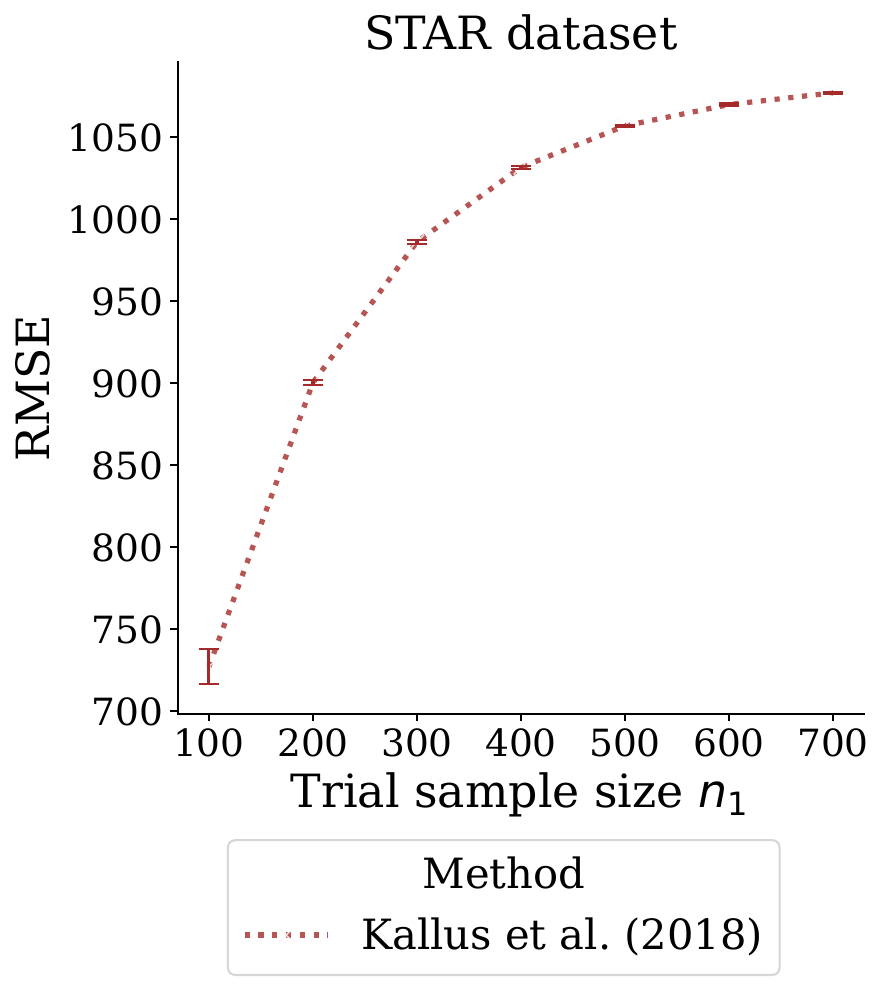}
        \caption{Rural schools as target population}
        \label{fig:app_star_rural_target_kallus}
    \end{subfigure}
    \caption{\small Separate plot for additive bias correction method of~\citet{kallus2018removing} because its large RMSE values make it difficult to display alongside the other methods. We evaluate the RMSE on the STAR dataset when increasing the trial sample size with a fixed external sample size of $n_0=1000$. We report the average RMSE and standard error over 200 repeated runs.}
    \label{fig:app_star_kallus}
\end{figure*}

We also include the performance of KSP, the additive bias correction method of~\citet{kallus2018removing}, in Figure~\ref{fig:app_star_kallus}. This method exhibits significantly higher RMSE than the other methods and, unusually, its performance worsens as the number of trial samples increases. While we cannot fully explain this behavior, one possible reason is that, unlike the other learners, this approach is not specifically designed for estimating a trial-specific CATE.

\newpage
\vspace{2cm}

\end{document}

%% file: head.tex
\usepackage[utf8]{inputenc} % allow utf-8 input
\usepackage[T1]{fontenc}    % use 8-bit T1 fonts
\usepackage{hyperref}       % hyperlinks
\usepackage{url}            % simple URL typesetting
\usepackage{booktabs}       % professional-quality tables
\usepackage{amsfonts}       % blackboard math symbols
\usepackage{nicefrac}       % compact symbols for 1/2, etc.
\usepackage{microtype}      % microtypography
\usepackage{xcolor}         % colors

\usepackage{amsmath,amssymb,dsfont,amsthm}       % blackboard math symbols
\usepackage{mathtools}
\usepackage{enumerate}
\usepackage{graphicx}
\usepackage{placeins}
\usepackage{float}
\usepackage{soul,color}
\usepackage{comment}
\usepackage{xcolor}
\usepackage{caption}
\usepackage{subcaption}
\usepackage{siunitx}
\usepackage{enumitem}
\usepackage{algorithm}
\usepackage{algorithmic}
\usepackage{mathrsfs}

\def\E{\mathbb{E}}
\def\V{\mathrm{Var}}

\DeclareMathOperator*{\argmin}{arg\,min}

\newcommand\indep{\protect\mathpalette{\protect\independenT}{\perp}}
\def\independenT#1#2{\mathrel{\rlap{$#1#2$}\mkern2mu{#1#2}}}

\newtheorem{thmlem}{Lemma}

\newtheorem{thmthm}{Theorem}

\newtheorem{thmcond}{Condition}
\newtheorem*{thmidasmp*}{Identifying assumptions}

\theoremstyle{definition}

\newtheorem{thmrem}{Remark}
\newtheorem*{proofsketch*}{Proof sketch}

\newcommand{\cate}{\tau}

\newcommand{\hattauDR}{\hat\tau_{DR}}
\newcommand{\hattauQR}{\hat\tau_{QR}}
\newcommand{\tauHat}{\hat\tau}

%% file: tables/n1_250_rmse.tex
\begin{tabular}{lllllll}
\toprule
External sample size & \multicolumn{2}{c}{100} & \multicolumn{2}{c}{1000} & \multicolumn{2}{c}{10000} \\
Condition 3 and 4 violated? & No & Yes & No & Yes & No & Yes \\
\midrule
Predict ATE & 0.31 (1e-3) & \textbf{0.31} (1e-3) & 0.31 (1e-3) & 0.31 (1e-3) & 0.31 (1e-3) & 0.31 (1e-3) \\
DR-learner & 0.28 (4e-3) & 0.32 (4e-3) & 0.28 (4e-3) & 0.32 (4e-3) & 0.27 (4e-3) & 0.32 (4e-3) \\
T-learner & 0.55 (3e-3) & 0.55 (3e-3) & 0.55(3e-3)& 0.55 (3e-3)& 0.55 (3e-3)& 0.55 (3e-3)\\
Pooled T-learner & 0.52 (2e-3) & 0.60 (3e-3) & 0.47 (9e-4) & 0.60 (1e-3) & 0.33 (7e-4) & 0.48 (1e-3) \\
CFACE {\hypersetup{hidelinks}\citep{asiaee2023leveraging}} & 0.34 (5e-3) & 0.36 (4e-3) & 0.24 (4e-3) & 0.30 (3e-3) & \textbf{0.19} (3e-3) & 0.28 (3e-3) \\
R-OSCAR {\hypersetup{hidelinks}\citep{asiaee2023improving}} & 0.60 (4e-03) & 0.60 (4e-03) & 0.52 (2e-03) & 0.58 (2e-03) & 0.37 (1e-03) & 0.41 (2e-03) \\
KSP {\hypersetup{hidelinks}\citep{kallus2018removing}} & 0.71 (1e-2) & 0.76 (1e-2) & 0.72 (1e-2) & 0.77 (1e-2) & 0.71 (1e-2) & 0.77 (1e-2) \\
QR-learner (ours) & \textbf{0.28} (4e-3) & 0.32 (4e-3) & \textbf{0.23} (3e-3) & \textbf{0.29} (3e-3) & \textbf{0.19} (3e-3) & \textbf{0.27} (3e-3) \\
Combined learner (ours) & 0.29 (4e-3) & 0.32 (4e-3) & \textbf{0.23} (3e-3) & \textbf{0.29} (4e-3) & \textbf{0.19} (3e-3) & \textbf{0.27} (3e-3) \\
\bottomrule
\end{tabular}

%% file: tables/n1_250_bias.tex
\begin{tabular}{lllllll}
\toprule
External sample size & \multicolumn{2}{c}{100} & \multicolumn{2}{c}{1000} & \multicolumn{2}{c}{10000} \\
Condition 3 and 4 violated? & No & Yes & No & Yes & No & Yes \\
\midrule
DR-learner & -0.00 (5e-03) & -0.00 (5e-03) & -0.00 (5e-03) & -0.01 (5e-03) & 0.00 (4e-03) & 0.01 (5e-03) \\
T-learner & 0.01 (4e-03) & 0.01 (5e-03) & 0.01 (5e-03) & 0.00 (5e-03) & 0.01 (4e-03) & 0.00 (5e-03) \\
Pooled T-learner & 0.06 (3e-03) & 0.12 (4e-03) & 0.05 (2e-03) & 0.20 (3e-03) & -0.03 (1e-03) & 0.11 (1e-03) \\
CFACE {\hypersetup{hidelinks}\citep{asiaee2023leveraging}} & 0.01 (5e-03) & 0.00 (5e-03) & 0.01 (4e-03) & -0.00 (5e-03) & 0.00 (3e-03) & 0.00 (4e-03) \\
R-OSCAR {\hypersetup{hidelinks}\citep{asiaee2023improving}} & 0.00 (6e-03) & 0.00 (5e-03) & 0.00 (4e-03) & 0.01 (5e-03) & -0.00 (3e-03) & 0.00 (4e-03) \\
KSP {\hypersetup{hidelinks}\citep{kallus2018removing}} & 0.01 (5e-03) & -0.01 (5e-03) & -0.01 (6e-03) & -0.01 (6e-03) & -0.01 (5e-03) & 0.01 (6e-03) \\
QR-learner (ours) & 0.01 (5e-03) & -0.00 (5e-03) & 0.01 (4e-03) & -0.00 (4e-03) & 0.00 (3e-03) & 0.00 (4e-03) \\
Combined learner (ours) & 0.00 (5e-03) & 0.00 (5e-03) & 0.00 (4e-03) & 0.00 (4e-03) & -0.00 (3e-03) & 0.00 (4e-03) \\
\bottomrule
\end{tabular}